\documentclass[10pt,journal,compsoc]{IEEEtran} 

\usepackage{bm}
\usepackage{amsmath}
\usepackage{epsf}
\usepackage{graphics}
\usepackage{ amssymb }
\usepackage[dvips]{graphicx}
\usepackage{mathtools}
\usepackage{epsfig}
\usepackage{cite}
\usepackage[linesnumbered,ruled,vlined]{algorithm2e}
\usepackage{colortbl}
\usepackage{color}
\usepackage{enumitem}
\usepackage{soul,xcolor}

\usepackage{bm}
\usepackage{amsmath}
\usepackage{epsf}
\usepackage{graphics}
\usepackage{ amssymb }
\usepackage[dvips]{graphicx}
\usepackage{epsfig}
\usepackage{cite}
\usepackage[linesnumbered,ruled,vlined]{algorithm2e}
\usepackage{graphicx}
\usepackage{epsfig}
\usepackage{latexsym}
\usepackage{amsfonts}
\usepackage{here}
\usepackage{rawfonts}
\usepackage[utf8]{inputenc}
\usepackage[english]{babel}
\usepackage{amsmath}
\usepackage{amsfonts}
\usepackage{amssymb}
\usepackage{color}
\usepackage{bm}
\usepackage{listings}
\usepackage{caption}
\usepackage{multicol}
\usepackage{tabularx,booktabs}
\newcolumntype{Y}{>{\centering\arraybackslash}X}
\makeatletter
\newcommand\notsotiny{\@setfontsize\notsotiny{6.31415}{7.1828}}
\makeatother
\usepackage{amssymb}
\usepackage{amsthm}
\usepackage{graphicx}
\usepackage{epstopdf}
\usepackage{listings}
\usepackage{float}
\usepackage{amsmath}
\usepackage{amssymb}
\usepackage{amsfonts}
\usepackage{epstopdf}

\usepackage{multirow}
\usepackage{amscd}
\usepackage{mathrsfs}
\usepackage{graphicx}
\usepackage{color}
\usepackage{url}
\usepackage{bm}
\usepackage{setspace}
\usepackage{footnote}
\usepackage{xcolor}
\DeclareMathOperator*{\argmax}{arg\,max}

\newtheorem{lemma}{Lemma}

\newtheorem{remark}{Remark}

\usepackage[]{algorithm2e}

\addto\captionsenglish{}
\usepackage{bbm}

\usepackage{multicol}
\usepackage{mathtools}

\usepackage{lipsum,graphicx,subcaption}

\usepackage{diagbox}
\usepackage{hyperref}
\usepackage[protrusion=true,expansion=true]{microtype}
\pdfoutput=1
\usepackage[font=footnotesize]{caption}
\allowdisplaybreaks[4]

\usepackage{graphicx}
\usepackage{grffile}
\usepackage{tabularx}
\newcounter{term}[section]

\renewcommand\theterm{\alph{term}}
\makeatletter
\newcommand{\vast}{\bBigg@{4}}

\newcommand{\Vast}{\bBigg@{5}}
\makeatother
\newcommand\semiHuge{\fontsize{22.7}{31.38}\selectfont}
\usepackage{soul,xcolor}

\usepackage[most]{tcolorbox}
\tcbset{colback=yellow!10!white, colframe=black!50!black, 
        highlight math style= {enhanced, 
            colframe=black,colback=red!10!white,boxsep=0pt}
        }
        \allowdisplaybreaks

\begin{document} 

\title{{\semiHuge DNN Partitioning, Task Offloading, and Resource
Allocation in Dynamic Vehicular Networks: A Lyapunov-Guided
Diffusion-Based Reinforcement Learning Approach}}
\author{Zhang Liu,~\IEEEmembership{Student Member,~IEEE}, Hongyang Du,~\IEEEmembership{Student Member,~IEEE}, Junzhe Lin,~\IEEEmembership{Student Member,~IEEE}, Zhibin Gao,~\IEEEmembership{Member,~IEEE}, Lianfen Huang,~\IEEEmembership{Member,~IEEE}, \\Seyyedali Hosseinalipour,~\IEEEmembership{Member,~IEEE}, and Dusit Niyato,~\IEEEmembership{Fellow,~IEEE}
\thanks{\emph{Z. Liu (zhangliu@stu.xmu.edu.cn), J. Lin (linjunzhe@stu.xmu.edu.cn), and L. Huang (lfhuang@xmu.edu.cn) are with the Department of Informatics and Communication Engineering, Xiamen University, Fujian, China 361102. H. Du (hongyang001@e.ntu.edu.sg), and D. Niyato (dniyato@ntu.edu.sg) are with the School of Computer Science and Engineering, Nanyang Technological University, Singapore 639798. Z. Gao (gaozhibin@jmu.edu.cn) is with the Navigation Institute, Jimei University, Xiamen, Fujian, China 361021. S. Hosseinalipour (alipour@buffalo.edu) is with the Department of Electrical Engineering, University at Buffalo-SUNY, Buffalo, NY 14260. (Corresponding author: Zhibin Gao).} }
}
\maketitle
\setulcolor{red}
\setul{red}{2pt}
\setstcolor{red}

\begin{abstract}
The rapid advancement of Artificial Intelligence (AI) has introduced Deep Neural Network (DNN)-based tasks to the ecosystem of vehicular networks. These tasks are often computation-intensive, requiring substantial computation resources, which are beyond the capability of a single vehicle. To address this challenge, Vehicular Edge Computing (VEC) has emerged as a solution, offering computing services for DNN-based tasks through resource pooling via Vehicle-to-Vehicle/Infrastructure (V2V/V2I) communications. In this paper, we formulate the problem of joint DNN partitioning, task offloading, and resource allocation in VEC as a dynamic long-term optimization. Our objective is to minimize the DNN-based task completion time while guaranteeing the system stability over time. To this end, we first leverage a Lyapunov optimization technique to decouple the original long-term optimization with stability constraints into a per-slot deterministic problem. Afterwards, we propose a Multi-Agent Diffusion-based Deep Reinforcement Learning (MAD2RL) algorithm, incorporating the innovative use of diffusion models to determine the optimal DNN partitioning and task offloading decisions. Furthermore, we integrate convex optimization techniques into MAD2RL as a subroutine to allocate computation resources, enhancing the learning efficiency. Through simulations under real-world movement traces of vehicles, we demonstrate the superior performance of our proposed algorithm compared to existing benchmark solutions.
\end{abstract}
\vspace{-3mm}
\section{Introduction} \label{sec:introduction}
\subsection{Background Research} \label{subsec:background}
\vspace{-.15mm}
Driven by recent breakthroughs in Artificial Intelligence (AI), Deep Neural Networks (DNNs) are now integrated into the ecosystem of vehicular networks, enabling a wide range of driving assistance applications. The DNN-based tasks, such as autonomous driving, simultaneous localization and mapping, and augmented reality navigation~\cite{9109630,8884164,9748064,10379539} hold immense potential to enhance the drivers' and passengers' safety and comfort. Taking autonomous driving as an example, vast quantities of data collected through smart cars' on-board camera and sensors are fed into pre-trained DNN models, such as VGG16~\cite{simonyan2015deep}. Subsequently, the VGG16 model processes this data to perform \emph{DNN inference}, which for example leads to the recognition of traffic signs.

However, the limited computation capability of a single vehicle hinders the smooth processing of these DNN-based tasks. For instance, image classification using ResNET152~\cite{7780459} necessitates approximately 22.6 billion and 33 trillion computation operations to process a single 224 × 224 image and videos with 30 fps, respectively~\cite{10141674}. In practice, smart vehicles like Tesla Model X are equipped with 8 cameras and 12 ultrasonic radars, continuously collecting large volumes data, calling for extensive computation demands. Given these considerations, providing low-latency computing services through effective task offloading polices for DNN-based tasks in vehicular networks is of paramount importance.

To this end, cloud computing offers a viable solution for handling computation-intensive tasks through offloading them from the edge of the network to cloud servers for processing. However, continuous data transfer from vehicles to cloud servers can impose a prohibitive traffic congestion on backhaul links, leading to a high communication latency~\cite{10100891}. As an alternative, Vehicular Edge Computing (VEC)~\cite{10048752,9720111} brings cloud computing capabilities closer to the edge of vehicular networks, facilitating real-time and low-latency processing of tasks. Specifically, VEC utilizes the computation resources of both moving vehicles and RoadSide Units (RSUs) equipped with edge servers, offering a scalable and flexible computing framework. Exploiting this technology, in this paper, we partition the computations of DNN-based tasks into two parts with one processed locally and the other offloaded to nearby \emph{edge nodes} (i.e., other vehicles or the RSU), aiming to accelerate the inference process.

\vspace{-3mm}
\subsection{Motivation and Main Challenges} \label{subsec:challenges}
\begin{figure}[!t]
\includegraphics[width=.48\textwidth]{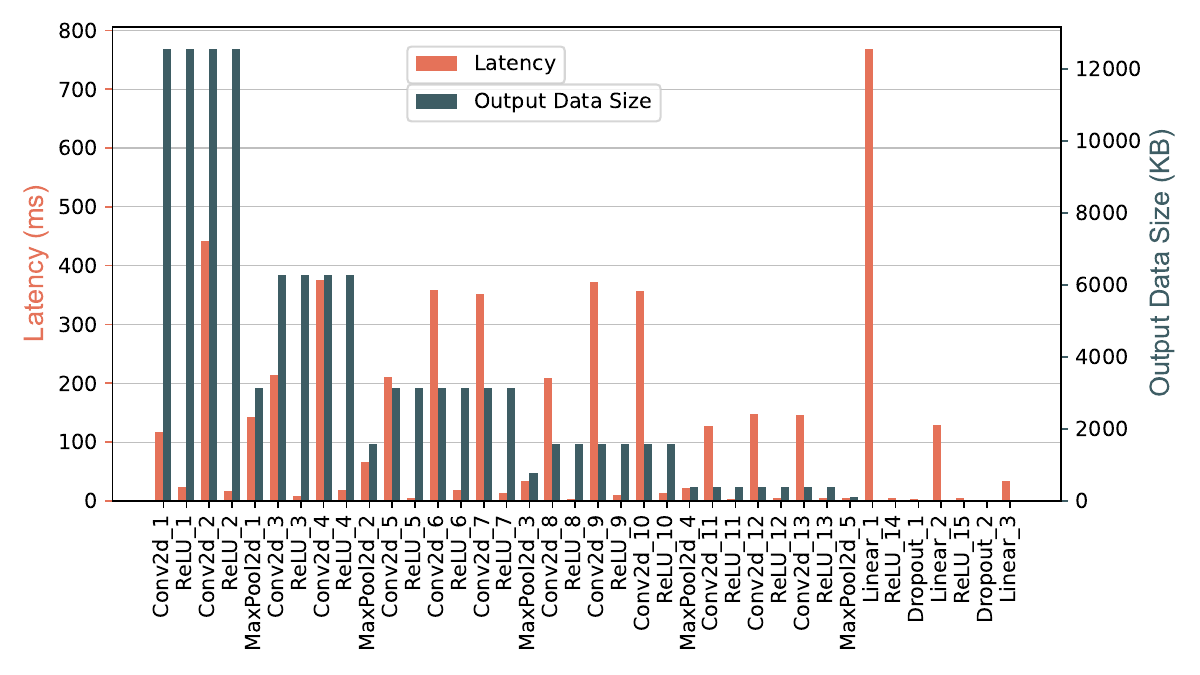}
\centering
\vspace{-1.5mm}
\caption{VGG16 layer-wise runtime and output data size.}
\label{fig:layer_demo}
\end{figure}

\vspace{-.15mm}
Although VEC is a promising solution to process DNN-based tasks, there still remains noteworthy issues that need to be addressed. Firstly, there exist non-uniformities in processing pipeline of DNN-based tasks. To demonstrate this, we conduct a pilot study on the layer-wise execution latency and the size of intermediate output data per layer for VGG16~\cite{simonyan2015deep}, utilizing GT-SRB dataset~\cite{8709983}, the results of which are illustrated in Fig.~\ref{fig:layer_demo}. By inspecting the results, a considerable heterogeneity in latency and output data size across different layers of VGG16 can be observed. This result further unveils the importance of choosing the optimal partitioning point for a DNN-based task, showcasing that arbitrary partitioning can fail to offload the computation-intensive part to edge nodes at a low transmission cost.

Secondly, when the computation of a DNN-based task is partitioned into two parts, with the intermediate data being offloaded to other edge nodes, the inherent mobility of vehicles introduces uncertainties into the offloading process due to the varying channel conditions. This in turn calls for the real-time decision making regarding the offloading/dispersion of the intermediate data across the edge nodes. However, existing approaches for offloading computation-intensive tasks, including heuristic-based~\cite{8854339},~\cite{8451923} and decomposition-oriented~\cite{8533343} search algorithms, often require extensive iterations to converge to a relatively good/stable solution. This makes them less practical for implementation over dynamic vehicular networks.

The third challenge lies in ensuring the stability of the system\footnote{In this paper, a stable system is characterized by its ability to ensure long-term equilibrium in task execution, which is dictated by the behavior of the task queues (to be detailed in Sec.~\ref{sec:system_model}). Specifically, these time-evolving task queues should not experience a consistent upward trend over time slots~\cite{10115012},~\cite{9449944}.} during the execution of DNN-based tasks. In particular, the majority of existing research on task offloading primarily aims at enhancing performance indicators, including task execution delay and energy consumption, as highlighted in ~\cite{10100891},\cite{8854339,8451923,8533343}. However, such studies often neglect the critical need for ensuring reliable system operations. This common oversight potentially jeopardizes the operational stability of the system. As a result, the complex dynamics involved in handling of DNN-based tasks in VEC, including the decisions on DNN partitioning, task offloading, and resource allocation across consecutive time slots are non-trivial to address, which is the main motivation behind this work. 

\vspace{-3.5mm}
\subsection{Summary of Contributions} \label{subsec:contributions}
\vspace{-.7mm}
In this paper, we propose a novel VEC-assisted DNN-based task execution framework, where vehicles strategically partition and offload their tasks to other edge nodes through V2V and V2I links. We formulate the
problem of joint DNN partitioning, task offloading, and resource allocation as a dynamic long-term optimization with the objective of minimizing the DNN-based task completion time while guaranteeing the system stability over time. Our main contributions are outlined as follows:
\begin{itemize}[leftmargin=4mm]
\item We formulate the long-term optimization problem of joint DNN partitioning, task offloading, and resource allocation in vehicular networks as a Mixed Integer NonLinear Program (MINLP). We discuss that the formulated MINLP is an NP-hard problem, and thus non-trivial to solve especially over large-scale dynamic vehicular networks.
\item To address this challenge, we first exploit a Lyapunov optimization technique. This strategy effectively decouples the long-term task completion time minimization with stability constraints into a per-slot deterministic problem. Following this transformation, we propose a Multi-Agent Diffusion-based Deep Reinforcement Learning (MAD2RL) algorithm to handle these per-slot problems. More importantly, our work is the first to integrate a diffusion model into the multi-agent reinforcement learning framework.
\item The MAD2RL algorithm incorporates the innovative use of a diffusion model -- initially invented for image generation -- to determine the optimal DNN partitioning and task offloading decisions. Specifically, the diffusion model operates by progressively reducing noise through a series of denoising steps, effectively extracting the optimal decisions from an initial state of Gaussian noise. Also, we integrate MAD2RL with convex optimization techniques for the allocation of computing resources, aiming to boost the convergence of its learning process.

\item We validate the effectiveness of our proposed algorithm through simulations on a real-world vehicular network obtained from OpenStreetMap~\cite{4653466}, utilizing the Simulation of Urban MObility (SUMO)~\cite{8569938} for integrating moving vehicles. We further conduct comparative analysis across real-world DNN models under different simulation settings, showcasing the superior performance of our approach.
\end{itemize}

\vspace{-5mm}
\subsection{Paper Organization} \label{subsec:organization}
\vspace{-.5mm}
The remainder of this paper is structured as follows: Sec.~\ref{sec:related_works} elaborates on the related works. Sec.~\ref{sec:system_model} details the system model. In Sec.~\ref{sec:probformu}, we formulate the problem of DNN partitioning, task offloading, and resource allocation in vehicular networks. Sec.~\ref{sec:Lyapunov} proposes Lyapunov optimization to handle the original problem. In Sec.~\ref{sec:DDPM}, we introduce the motivation of adopting the diffusion model. Our proposed MAD2RL algorithm is presented in Sec.~\ref{sec:D2RL}. Simulation results are detailed in Sec.~\ref{sec:simulation}, followed by concluding the paper in Sec.~\ref{sec:conclusion}.

\vspace{-3mm}
\section{Related Works} \label{sec:related_works}
\vspace{-.7mm}
Henceforth, we summarize contributions of the related works and discuss the distinctions between our methodology and prior research.

\vspace{-3mm}
\subsection{DNN Inference and Edge Computing}\label{subsec:DNNinference}
\vspace{-.5mm}
Researchers in~\cite{9384272} introduced an device-edge-cloud orchestration architecture, optimizing the assignment of inference tasks and deployment of DNN models for maximizing the DNN task inference accuracy. Researchers in~\cite{9170818} investigated collaborative DNN inference within industrial networks, addressing the sampling rate adaptation for sensing data, inference task offloading, and the allocation of edge computing resources. Researchers in~\cite{9542866} focused on a joint strategy for task partitioning and offloading for DNN-based tasks in mobile edge computing networks to minimize the computation cost incurred on the devices. 

Despite their notable contributions, these studies overlook the unique characteristics of VEC systems, such as the mobility of vehicles and varying channel conditions. Besides, the aforementioned works neglect the use of computation resources of nearby vehicles through V2V links, rendering their approaches less applicable to the VEC paradigm. 

\vspace{-3mm}
\subsection{DRL for Resource Management}\label{subsec:DRL}
\vspace{-.5mm}
In response to addressing the complexities of dynamic computing environments, recent studies have explored learning-based approaches for a variety of networking problems, with Deep Reinforcement Learning (DRL) being a notable example~\cite{9234039,10056271,9552188}. Generally speaking, DRL employs DNNs to learn the relationship between state space of the problems (e.g., channel conditions and computational resource availability) and their action space (e.g., task offloading decisions) without requiring prior knowledge about the environment characteristics (e.g., vehicle mobility patterns). Researchers in~\cite{9720111} proposed a double deep Q-learning DRL algorithm to motivate vehicles to share their computing resources while guaranteeing the reliability of resource allocation in VEC. Researchers in~\cite{10056271} utilized a multi-agent soft actor-critic DRL approach with an attention mechanism to learn the strategic trading for each vehicle in a vehicular fog computing system. Researchers in~\cite{9552188} employed a joint offloading and resource allocation algorithm based on the Multi-Agent Deep Deterministic Policy Gradient (MADDPG) DRL algorithm to decrease the vehicle’s energy cost for executing tasks while increasing the revenue of the vehicle for processing tasks in a VEC network. 

It is worth mentioning that the primary focus of DRL-based methodologies in aforementioned works is on enhancing the computation performance, failing to guarantee the system stability over time, which is vital in real-world networks. Moreover, the aforementioned studies overlook the importance of scheduling DNN-based tasks, which has been heightened by recent breakthroughs in AI, further motivating this work.

\vspace{-3mm}
\subsection{Lyapunov Optimization for System Stability}\label{subsec:Lyapunov}
Lyapunov optimization has been recognized as a promising approach for offering the dual benefits of maximizing the system utility while ensuring the stability of system operations. Researchers in~\cite{10115012} introduced a Lyapunov-based DDPG method for the joint optimization of computation task distribution and radio resource allocation within vehicular networks. Researchers in~\cite{9449944} developed a Lyapunov-guided DRL-based algorithm for making task offloading and system resource allocation decisions in a dynamic edge computing environment. Researchers in~\cite{10102429} employed a Lyapunov-guided DRL resource management strategy to reduce the average power consumption across the system. Researchers in~\cite{9964433} leveraged Lyapunov optimization in conjunction with a deep Q-network to select optimal task offloading actions that ensure long-term task queue stability. 

However, the aforementioned works merely focus on bit-stream computation tasks, overlooking the intricacies involved in DNN partitioning. In this work, we unveil an end-to-end framework, entailing the combination of Lyapunov optimization, diffusion models, and DRL techniques for efficient execution of DNN-based tasks in dynamic VEC environments.

\vspace{-3mm}
\subsection{Footprints of Diffusion Models in Optimization}\label{subsec:diffusion}
\vspace{-.5mm}
Recently, diffusion models~\cite{ho2020denoising} have emerged as powerful deep generative tools in machine learning, gaining prominence especially in image and video generation fields. These models often operate through incrementally introducing noise to an original image until it becomes indistinguishable from Gaussian noise (the forward process). Subsequently, they learn to invert this diffusion process, restoring the original image (the reverse process). The groundbreaking innovation of applying diffusion models in learning-based methods was achieved by works such as~\cite{wang2023diffusion}, which introduced a diffusion Q-learning method for behavior cloning and policy regularization within offline reinforcement learning frameworks. Researchers in~\cite{du2023diffusionbased} crafted a diffusion model-based DRL method for selecting optimal service providers for AI content creation in Metaverse. Researchers in~\cite{10158526} introduced an innovative diffusion model-based algorithm for AI-generated contract design, facilitating information sharing in semantic communications. 

We contribute to this literature via introducing a new application of diffusion-based models in DNN-task processing, entailing a Lyapunov-guided multi-agent diffusion-based deep reinforcement learning approach for VEC that can minimize the DNN-based task completion time while guaranteeing the system stability over time.

\vspace{-3mm}
\section{System Model} \label{sec:system_model}
\vspace{-.7mm}
Henceforth, we first present the AI-powered vehicular network model and introduce the DNN partitioning model. Then, we formulate the communication model, followed by the computing model.

\vspace{-1.5mm}
\subsection{AI-Powered Vehicular Network Model}\label{subsec:vehicular_networks}
\begin{figure}[t!]
\includegraphics[width=.48\textwidth]{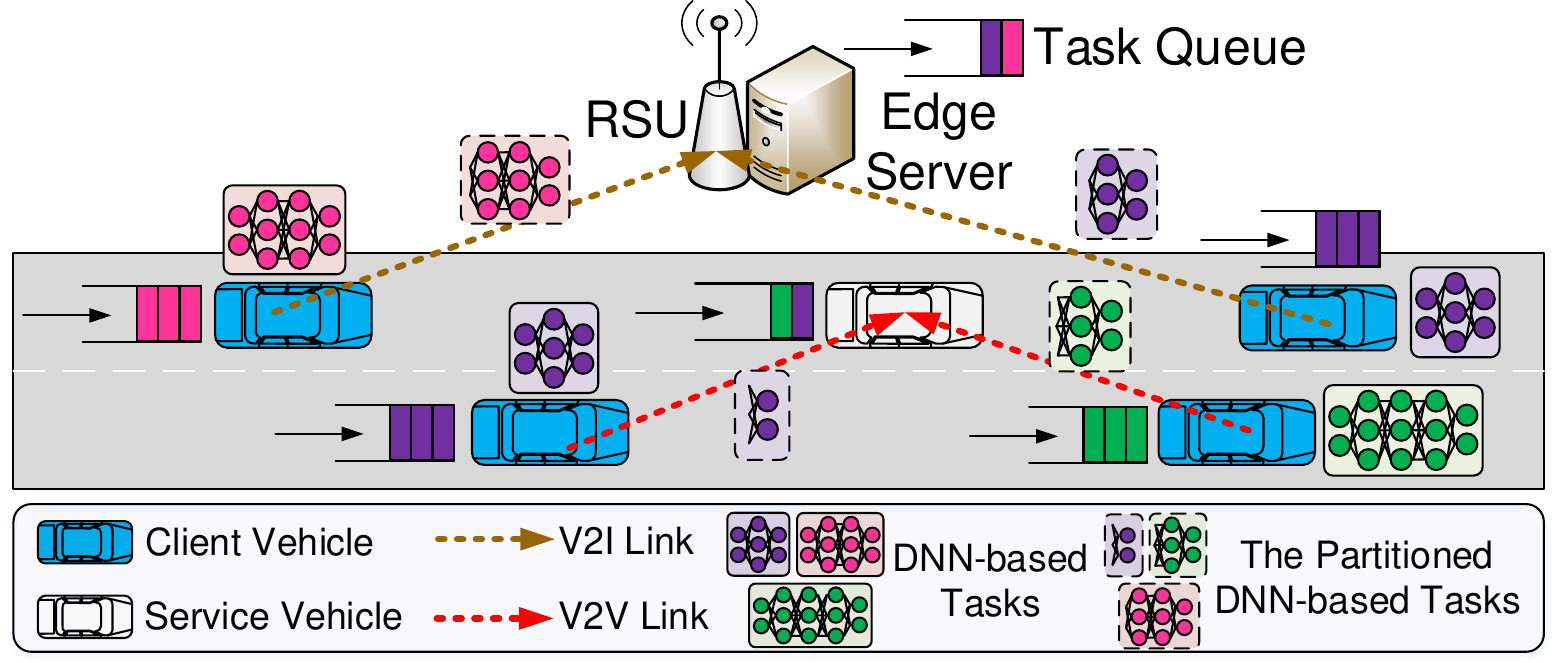}
\centering
\vspace{-2mm}
\caption{A schematic of VEC-assisted DNN-based task partitioning and offloading.}
\label{fig:Vehicular_Networks}
\vspace{-1mm}
\end{figure}

\vspace{-.15mm}
We consider an AI-powered vehicular network, in which vehicles possessing DNN-based tasks move on an unidirectional highway. Similar to~\cite{10100891},~\cite{8543658}, and~\cite{8345717}, our analysis are focused on the decision making and network control at an arbitrary RSU as illustrated in Fig.~\ref{fig:Vehicular_Networks}. We adopt a discrete time slot representation of the system, where time slots are collected via the set $\begin{aligned}\mathcal{T}=\{1,. . ., T\}\end{aligned}$, and the duration of each time slot is $\tau$. The vehicles with DNN-based tasks are considered as \emph{Client Vehicles} (CVs), collected by the set $\begin{aligned}\mathcal{I}=\{1,. . .,I\}\end{aligned}$, while the RSU and the rest of vehicles, referred to as \emph{Service Vehicles} (SVs) are considered as edge nodes, the set of which is represented as $\begin{aligned}\mathcal{J}=\{1,2,. . .,J\}\end{aligned}$, where index $j=1$ is for the RSU and indices $j\in\{2,. . ., J\}$ are used for SVs. We assume that each CV $i$ can generate a DNN-based task $b_i$ at the beginning of each time slot $t$.

Additionally, two key assumptions are made in our following system model: (i) The channel condition are stable during one time slot~\cite{8057297}; (ii) The DNN models are well-trained and pre-loaded in each vehicle and the RSU without the need for further deployment~\cite{10105846}.

\vspace{-1mm}

\vspace{-5mm}
\subsection{DNN Partitioning Model}\label{subsec:partitioning_model}
We consider a VEC with $K$ types of DNN models, collected by the set $\mathcal{K}=\{1,. . .,K\}$ (e.g., ResNet18~\cite{7780459} and VGG16~\cite{simonyan2015deep}). Let $\mathcal{L}_{k} = \{1,. . ., L_{k}\}$ denote the set of layers of $k$th type DNN-based task, which includes convolution/pooling layers for feature extraction, followed by a series of fully connected layers for classification. Specifically, let $L^{\mathsf{con}}_{k}$ denote the index of the last convolution/pooling layer in $\mathcal{L}_{k}$. Each convolution/pooling layer $l_{k} \ (1\leq l_{k}\leq L^{\mathsf{con}}_{k})$ is characterized by a tuple $(H_{l_k}, W_{l_k}, C^{\mathsf{in}}_{l_k},C^{\mathsf{out}}_{l_k},ker_{l_k})$, where $(H_{l_k}, W_{l_k}, C^{\mathsf{in}}_{l_k})$ capture the value of input height, width and channel, and $(C^{\mathsf{out}}_{l_k},ker_{l_k})$ represent the value of output channel, and kernel size of layer $l_k$, respectively. Also, each fully connected layer $l_{k} \ (L^{\mathsf{con}}_{k} <l_{k}\leq L_{k})$ is represented by a tuple $(U^{\mathsf{in}}_{l_k},U^{\mathsf{out}}_{l_k})$, where $U^{\mathsf{in}}_{l_k}$ and $U^{\mathsf{out}}_{l_k}$ denote the size of the unidimensional input, and output of layer $l_{k}$, respectively.

Let the set of CVs possessing the $k$th type DNN model be denoted by $\mathcal{I}_k$\footnote{Although there are several types of DNN models, such as AlexNet~\cite{10.1145/3065386}, ResNet18~\cite{7780459}, and VGG16~\cite{simonyan2015deep}, each CV $i$ is assumed to generate a single type DNN-based task~\cite{9170818}.}, where $\mathcal{I}=\bigcup_{k\in \mathcal{K}}\mathcal{I}_k$. We model the DNN partitioning decision of CV $i$ at time slot $t$ through introducing an integer variable $\varphi_{i}(t)\in \{1,. . .,L_{k}+1\}, i \in \mathcal{I}_k$. According to this controllable variable, the DNN-based task $b_i$ is divided into two parts: the first part of layers 1 to $\varphi_{i}(t)-1$ are processed locally on CV $i$; while the second part of layers $\varphi_{i}(t)$ to $L_{k}$ are processed by an edge node (i.e., either the RSU or one of the SVs) after the transmission of input data of layer $\varphi_{i}(t)$ to that edge node. In particular, the whole DNN-based task $b_i$ will be locally computed at CV $i$ if $\varphi_{i}(t)=L_{k}+1$ or completely offloaded to an edge node if $\varphi_{i}(t)=1$.

\vspace{-5mm}
\subsection{Communication Model} \label{subsec:comm_model}
In the VEC, we consider Vehicle-to-Infrastructure (V2I) and
Vehicle-to-Vehicle (V2V) communication models concurrently. Since the communications between CVs, SVs, and the RSU can be facilitated using the Orthogonal Frequency Division Multiple Access (OFDMA) protocol, which effectively reduces the overlaps between the signals, we disregard the interference in the following discussions~\cite{10115012},~\cite{10105846}. Hereafter, we formulate the V2I and V2V communication models and then discuss the communication latency.
\vspace{-3mm}
\subsubsection{V2I Communications} The (maximum) transmission rate from CV $i$ to the RSU at time slot $t$ is given
by 
\vspace{-1.5mm}
\begin{equation}\label{eq:v2i_transrate}
R^{\mathsf{rsu}}_i(t)= \mathcal{B}\text{log}_2\Big(1+\frac{p_ih^{\mathsf{rsu}}_{i}(t)}{\sigma^2} \Big),
\vspace{-.5mm}
\end{equation}
where $\mathcal{B}$ (in Hz) denotes the (sub-)channel bandwidth,  $p_i$ (in W) represents the transmit power of CV $i$, and $\sigma^2$ (in W) is the noise power. Additionally, $h^{\mathsf{rsu}}_{i}(t)= g^{\mathsf{rsu}}_{i}(t)|u^{\mathsf{rsu}}_{i}(t)|^2$ (in dB)~\cite{9562522} is the V2I channel gain, which captures both large-scale and small-scale fading. Specifically, $g^{\mathsf{rsu}}_{i}(t)=-38.4-21.0\text{log}_{10}dis^{\mathsf{rsu}}_i(t)$~\cite{8944302} represents the V2I path loss, where $dis^{\mathsf{rsu}}_i(t)$ (in meters) indicates the Euclidean distance between CV $i$ and the RSU at time slot $t$, and $u^{\mathsf{rsu}}_{i}(t) \sim \mathcal{CN}(0,1)$ is the small-scale fading component, varying i.i.d. over time.  
\vspace{-2mm}
\subsubsection{V2V Communications} Similarly, focusing on the V2V links, the (maximum) V2V transmission rate from CV $i$ to SV $j$ at time slot $t$, denoted by $R^{\mathsf{veh}}_{i,j}(t)$, can be given by
\vspace{-1.5mm}
\begin{equation}\label{eq:v2v_transrate}
R^{\mathsf{veh}}_{i,j}(t)= \mathcal{B}\text{log}_2\Big(1+ \frac{p_ih^{\mathsf{veh}}_{i,j}(t)}{\sigma^2} \Big),
\vspace{-.5mm}
\end{equation}
where $h^{\mathsf{veh}}_{i,j}(t)=g^{\mathsf{veh}}_{i,j}(t)|u^{\mathsf{veh}}_{i,j}(t)|^2$ (in dB)~\cite{9562522} is the V2V channel gain between CV $i$ and SV $j$ at time slot $t$, which captures both large-scale and small-scale fading. Specifically, $g^{\mathsf{veh}}_{i,j}(t)=-44.23-16.7\text{log}_{10}dis^{\mathsf{veh}}_{i,j}(t)$~\cite{8944302} represents the V2V path loss, and $u^{\mathsf{veh}}_{i,j}(t) \sim \mathcal{CN}(0,1)$ is the small-scale fading component, varying i.i.d. over time.
\vspace{-2mm}
\subsubsection{Communication Latency} Upon partitioning a DNN-based task into two parts, the corresponding input data size of layer $l_{k}$ for transmitting the intermediate in-layer data is given by~\cite{10141674}
\vspace{-1.5mm}
\begin{equation}\label{eq:datasize_layer}
D_{l_{k}}= \left\{ \begin{array}{l}
			H_{l_{k}}W_{l_{k}}C^{\mathsf{in}}_{l_{k}}\varrho, \quad\quad 1\leq l_{k}\leq L_{k}^{\mathsf{con}}\\
			U^{\mathsf{in}}_{l_{k}}\varrho, \quad\quad\quad\quad\quad L_{k}^{\mathsf{con}}< l_{k} \leq L_{k}
		\end{array} \right. ,
\vspace{-.5mm}
\end{equation}
where $\varrho$ is the memory footprint for a unit data. Consequently, for DNN-based task $b_i$, when the DNN partitioning is made at layer $\varphi_{i}(t)$, the time required for intermediate data transmission from CV $i$ to the RSU can be calculated as
\vspace{-1.5mm}
\begin{equation}\label{eq:v2i_offloadtime}
d^{\mathsf{rsu,tra}}_{i}(t)= \left\{ \begin{array}{l}
            0, \quad\quad\quad\quad \varphi_{i}(t)=L_{k}+1 \\
			\xi_{i,1}(t)\frac{D_{\varphi_{i}(t)}}{R^{\mathsf{rsu}}_i(t)}, \quad \text{otherwise}
		\end{array} \right.\!\!\!\!,\ i \in \mathcal{I}_k,
\vspace{-.5mm}
\end{equation}
where $\varphi_{i}(t)=L_{k}+1$ means that the whole DNN-based task $b_i$ will be locally processed without data transmission, and $\xi_{i,j}(t)$ denotes the offloading decision of DNN-based task $b_i$ at time slot $t$. Specifically, $\xi_{i,j}(t)=1$ indicates that the remaining layers are offloaded to the RSU ($j=1$) or SV $j$ ($\forall j\in \mathcal{J}\setminus \{1\}$) for processing; $\xi_{i,j}(t)=0$ otherwise. Similarly, the time required for intermediate data transmission from CV $i$ to SV $j$, denoted by $d^{\mathsf{veh,tra}}_{i,j}(t)$, is given by
\vspace{-1.5mm}
\begin{equation}\label{eq:v2v_offloadtime}
d^{\mathsf{veh,tra}}_{i,j}(t)= \left\{ \begin{array}{l}
            0, \quad\quad\quad\quad \varphi_{i}(t)=L_{k}+1 \\
			\xi_{i,j}(t)\frac{D_{\varphi_i(t)}}{R^{\mathsf{veh}}_{i,j}(t)}, \quad \text{otherwise}
		\end{array} \right.\!\!\!\!,\ i \in \mathcal{I}_k.
\vspace{-.5mm}
\end{equation}
Since the transmission time of returning the inference result is much shorter than that of DNN inference and intermediate data transmission~\cite{10373167}, we neglect the feedback time.

\vspace{-3mm}
\subsection{Computing Model}  \label{subsec:comp_model}  
\vspace{-.5mm}
To quantify the processing intensity of each DNN-based task, we denote the computation workload of layer $l_{k}$ by $B_{l_{k}}$ (in the number of floating-point operations), which is given by~\cite{10141674} 
\vspace{-1.5mm}
\begin{equation}\label{eq:workload_layer}
B_{l_{k}}\!=\! \left\{ \begin{array}{l} 2
			H_{l_{k}}W_{l_{k}}(C^{\mathsf{in}}_{l_{k}}ker_{l_{k}}^2+1)C^{\mathsf{out}}_{l_{k}},\ 1\leq l_{k}\leq L_{k}^{\mathsf{con}} \\
			(2U^{\mathsf{in}}_{l_{k}}-1)U^{\mathsf{out}}_{l_{k}}, \ L_{k}^{\mathsf{con}}< l_{k} \leq L_{k}
		\end{array} \right.\!\!\!\!.
\vspace{-.5mm}
\end{equation}
We next divide DNN computing models into three parts: \emph{1) local processing}; \emph{2) edge processing at the RSU}; \emph{3) edge processing at the SVs}, and discuss them in order.
\vspace{-3mm}
\subsubsection{Local Processing} \label{subsubsec:local_processing}  
At time slot $t$, the local processing latency of the first segment of layers $l_k=1$ to $\varphi_{i}(t)-1$ of DNN-based task $b_i$, denoted by $d^{\mathsf{loc}}_{i}(t)$, is given by
\vspace{-1.5mm}
\begin{equation}\label{eq:local_delay}
d^{\mathsf{loc}}_{i}(t)\!=\! \left\{ \begin{array}{l}
            0, \quad\quad\quad\quad\quad\quad\quad\quad \varphi_{i}(t)=1\\
			\frac{Q_i^{\mathsf{loc}}(t)+\sum_{l_{k}=1}^{\varphi_{i}(t)-1}B_{l_{k}}}{f^{\mathsf{loc}}_i}, \ \text{otherwise}
		\end{array} \right.\!\!\!\!,\ i \in \mathcal{I}_k,
\vspace{-.5mm}
\end{equation}
where $\varphi_{i}(t)=1$ means that the entire DNN-based task $b_i$ is offloaded to a destination edge node without local computing, $f^{\mathsf{loc}}_i$ (in CPU cycle frequency) is the computation capacity of CV $i$, and $Q_i^{\mathsf{loc}}(t)$ is the backlogged computations in the local computing queue of CV $i$ at time slot $t$, which can be formulated through the following update rule:
\vspace{-1.5mm}
\begin{align}\label{eq:local_queue}
 &Q_i^{\mathsf{loc}}(t+1)= \max \Big\{Q_i^{\mathsf{loc}}(t)- \overbrace{f^{\mathsf{loc}}_i\tau}^{(\text{I})}\nonumber\\
 &+\overbrace{\mathbb{I}_{\{\varphi_{i}(t)\neq1\}}\Big(\sum_{l_{k}=1}^{\varphi_{i}(t)-1}B_{l_{k}}\Big)}^{(\text{II})},0 \Big\}, \ i\in \mathcal{I}_k.
\vspace{-.5mm}
\end{align}
Here, the term (I) indicates the computations that are executed within time slot duration $\tau$, and the term (II) signifies the newly arrived computations affected by the DNN partitioning decision $\varphi_{i}(t)$, where $\mathbb{I}_{\{\cdot\}}$ is an indicator function with $\mathbb{I}_{\{\cdot\}}=1$, when the condition in the argument is met; otherwise $\mathbb{I}_{\{\cdot\}}=0$.
\vspace{-3mm}
\subsubsection{Edge Processing at the RSU} \label{subsubsec:RSU_processing} 
We assume that the computing resources and the task queue located in the RSU are allocated among different DNN model types, which can be realized via containerization techniques~\cite{7036275}, indicating that the CVs with the same DNN type will be provided with the same computation resources at the RSU. When layers $l_{k}=\varphi_{i}(t)$ to $L_{k}$ are processed on the RSU, besides the V2I intermediate data transmission time in~\eqref{eq:v2i_offloadtime}, the task completion time further consists of task processing time, and waiting time, which are formalized below.
\begin{itemize}[leftmargin=4mm]
\item Processing time: Similar to~\eqref{eq:local_delay}, the processing time of DNN-based task $b_i, i \in \mathcal{I}_k$, at the RSU at time slot $t$, denoted by $d^{\mathsf{rsu,pro}}_{i}(t)$, can be calculated as
\vspace{-1.5mm}
\begin{equation}\label{eq:rsu_compdelay}
\hspace{-1mm} d^{\mathsf{rsu,pro}}_{i}(t)\!=\! \left\{ \begin{array}{l} 
0, \quad\quad\quad\quad\quad \varphi_{i}(t)=L_{k}+1 \\
\xi_{i,1}(t)\frac{Q^{\mathsf{rsu}}_{k}(t)+\sum_{l_{k}=\varphi_{i}(t)}^{L_{k}}B_{l_{k}}}{f_{k}^{\mathsf{rsu}}(t)}, \ \text{otherwise}
		\end{array} \right.\!\!\! , 
\vspace{-.5mm}
\end{equation}
where $\varphi_{i}(t)=L_{k}+1$ implies that the entire DNN-based task $b_i$ will be locally processed, $f_{k}^{\mathsf{rsu}}(t)$ is the computation resources of the RSU allocated to the $k$th type DNN model at time slot $t$, and $Q^{\mathsf{rsu}}_k(t)$ indicates the backlogged computations at the task queue of the RSU for $k$th type DNN model at time slot $t$, which can be represented via the following update rule
\vspace{-1.5mm}
\begin{align}\label{eq:rsu_queue}
    &Q^{\mathsf{rsu}}_k(t+1)=\max \Big\{Q_k^{\mathsf{rsu}}(t)-f^{\mathsf{rsu}}_k(t)\tau \nonumber\\
    &+\overbrace{\sum_{i \in \mathcal{I}_k}\xi_{i,1}(t)\mathbb{I}_{\{\varphi_{i}(t)\neq L_{k}+1\}}\Big(\sum_{l_{k}=\varphi_{i}(t)}^{L_{k}}B_{l_{k}}\Big)}^{(\text{I})},0 \Big\},
    \vspace{-.5mm}
\end{align}
where $Q^{\mathsf{rsu}}_k(1)=0, \forall k \in \mathcal{K}$, and the term (I) represents the newly arrived computations from all CV $i\in \mathcal{I}_k$ that chooses to offload their task to the RSU at time slot $t$.

\item Waiting time: The task waiting time, denoted by $d^{\mathsf{rsu,wait}}_{i}(t)$ consists of the average sojourn time among all newly arrived tasks at the RSU until DNN-based task $b_i, \ i \in \mathcal{I}_k$ is processed, which can be calculated as follows: 
\vspace{-1.5mm}
\begin{align}\label{eq:rsu_newtaskdelay}
    &d^{\mathsf{rsu,wait}}_{i}(t)=\nonumber\\
    &\frac{\sum_{i^{\prime}\in \mathcal{I}_k \setminus \{i\}}\xi_{i^{\prime},1}(t)\mathbb{I}_{\{\varphi_{i^{\prime}}(t)\neq L_{k}+1\}}\Big(\sum_{l_{k}=\varphi_{i^{\prime}}(t)}^{L_{k}}B_{l_{k}}\Big)}{2f^{\mathsf{rsu}}_k(t)}.
    \vspace{-.5mm}
\end{align}
\end{itemize}
In~\eqref{eq:rsu_newtaskdelay}, the division by 2 in the denominator is obtained using the technique used in~\cite{9170818}. As a result, the total completion time of DNN-based task $b_i$ on the RSU, denoted by $d^{\mathsf{rsu}}_i(t)$, can be calculated as follows: 
\vspace{-1.5mm}
\begin{align}\label{eq:total_rsudelay}
    d^{\mathsf{rsu}}_i(t)= d^{\mathsf{rsu,tra}}_i(t)+d^{\mathsf{rsu,pro}}_i(t)+d^{\mathsf{rsu,wait}}_i(t).
    \vspace{-.5mm}
\end{align}

\vspace{-3mm}
\subsubsection{Edge Processing at the SVs} \label{subsubsec:SV_processing} 
When remaining layers $l_k=\varphi_i(t)$ to $L_k$ are offloaded and processed on a nearby SV, besides the V2V intermediate data transmission time in~\eqref{eq:v2v_offloadtime}, the processing time of DNN-based task $b_i, i \in \mathcal{I}_k$ on SV $j \in \mathcal{J} \setminus \{1\}$ at time slot $t$, denoted by $d^{\mathsf{veh,pro}}_{i,j}(t)$, is given by
\vspace{-1.5mm}
\begin{align}\label{eq:sv_comp_delay}
    \hspace{-1mm}d^{\mathsf{veh,pro}}_{i,j}(t)\!=\!\!\left\{ \begin{array}{l}      
            0, \quad\quad\quad\quad\quad \varphi_{i}(t)=L_{k}+1 \\
			\xi_{i,j}(t)\frac{Q^{\mathsf{veh}}_j(t)+\sum_{l_k=\varphi_{i}(t)}^{L_{k}}B_{l_k}}{f^{\mathsf{veh}}_{j}}, \ \text{otherwise}
		\end{array} \right.\!\!\!\!, 
\vspace{-.5mm}
\end{align}
where $f^{\mathsf{veh}}_{j}$ is the computation capability of SV $j$, and $Q^{\mathsf{veh}}_j(t)$ represents the backlogged computations in the task queue of SV $j$, which can be represented via the following update rule
\vspace{-1.5mm}
\begin{align}\label{eq:sv_queue}
    &Q^{\mathsf{veh}}_j(t+1)=\max \Big\{Q^{\mathsf{veh}}_j(t)-f^{\mathsf{veh}}_j\tau \nonumber\\
    &+\sum_{k \in \mathcal{K}}\sum_{i \in \mathcal{I}_k}\xi_{i,j}(t)\mathbb{I}_{\{\varphi_{i}(t)\neq L_{k}+1\}}\Big(\sum_{l_{k}=\varphi_{i}(t)}^{L_{k}}B_{l_{k}}\Big),0 \Big\},
    \vspace{-.5mm}
\end{align}
with $Q_j^{\mathsf{veh}}(1)=0$. Similar to~\eqref{eq:rsu_newtaskdelay}, the average sojourn time among all newly arrived tasks at SV $j$ until the DNN-based task $b_i$ is processed, denoted by $d^{\mathsf{veh,wait}}_{i,j}(t)$, where $i^{\prime} \in\mathcal{I}_k$, is given by
\vspace{-1.5mm}
\begin{align}\label{eq:sv_newtaskdelay}
    &d^{\mathsf{veh,wait}}_{i,j}(t)=\nonumber\\
    &\frac{\sum_{i^{\prime}\in \mathcal{I}\setminus \{i\}}\xi_{i^{\prime},j}(t)\mathbb{I}_{\{\varphi_{i^{\prime}}(t)\neq L_{k}+1\}}\Big(\sum_{l_{k}=\varphi_{i^{\prime}}(t)}^{L_{k}}B_{l_{k}}\Big)}{2f^{\mathsf{veh}}_j}.
    \vspace{-.5mm}
\end{align}

Finally, the completion time of DNN-based task $b_i$ on SV $j$ can be calculated as
\vspace{-1.5mm}
\begin{align}\label{eq:sv_totaldelay}
    d^{\mathsf{veh}}_{i,j}(t)= d^{\mathsf{veh,tra}}_{i,j}(t)+d^{\mathsf{veh,pro}}_{i,j}(t)+d^{\mathsf{veh,wait}}_{i,j}(t).
    \vspace{-.5mm}
\end{align}

In summary, taking the aforementioned three processing forms into consideration, the total delay for completing DNN-based task $b_i$ at time slot $t$ can be calculated by
\vspace{-1.5mm}
\begin{align}\label{eq:total_delay}
d_{i}(t)= d^{\mathsf{loc}}_i(t)+d^{\mathsf{rsu}}_{i}(t)+d^{\mathsf{veh}}_{i,j}(t).
\vspace{-.5mm}
\end{align} 

\begin{table*}[!t]
\vspace{-3mm}
 \begin{equation}\label{eq:constant_value}
 \hspace{-13mm}
 \begin{aligned}
\chi&=\frac{1}{2}\sum_{i\in \mathcal{I}}\Big(\sum_{l_k=1}^{L_k}B_{l_k}\Big)^2+\frac{1}{2}\sum_{i \in \mathcal{I}}(f^{\mathsf{loc}}_i\tau)^2+ \frac{1}{2}\sum_{k \in\mathcal{K}}\Big( \sum_{i \in \mathcal{I}_k} \sum_{l_{k}=1}^{L_{k}}B_{l_{k}}\Big)^2 + \frac{1}{2}\sum_{k \in\mathcal{K}}(f^{\mathsf{rsu}}_k(t)\tau)^2\\
&+\frac{1}{2}\sum_{j\in \mathcal{J}\setminus\{1\}}\Big(\sum_{k \in \mathcal{K}}\sum_{i \in \mathcal{I}_k}\sum_{l_{k}=1}^{L_{k}}B_{l_{k}}\Big)^2 +\frac{1}{2}\sum_{j\in \mathcal{J}\setminus\{1\}}(f^{\mathsf{veh}}_j\tau)^2 
\end{aligned}
 \end{equation}
 \hrule
 \vspace{-3mm}
\end{table*}

\vspace{-3mm}
\section{Problem Formulation} \label{sec:probformu}
\vspace{-.7mm}
We next formulate the problem of joint DNN partitioning, task offloading, and resource allocation as a dynamic long-term optimization, aiming to minimize the DNN-based task completion time across all CVs while guaranteeing the system stability over time. Mathematically, we formulate this problem as optimization problem $(\mathcal{P}_1)$ given below:
\vspace{-1.5mm}
\begin{align}
     &(\mathcal{P}_1): \min_{\{\bm{\varphi}(t),\bm{\xi}(t),\bm{F}(t)\}_{t\in \mathcal{T}}}\lim_{T \rightarrow +\infty } \frac{1}{T}\sum_{t\in\mathcal{T}}\sum_{i\in\mathcal{I}}\mathbb{E}[d_i(t)]\hspace{-40mm}\label{eq:problem_1} \\[-.45em]
     & \textrm{s.t.}\nonumber\\ 
     & \lim_{T\rightarrow +\infty}\frac{1}{T}\sum_{t \in \mathcal{T}}\mathbb{E}\big[Q^{\mathsf{loc}}_i(t)\big]<\infty, \ \forall  i\in \mathcal{I}, \label{eq:Qloc_constraint}\\[-.2em]
     & \lim_{T\rightarrow +\infty}\frac{1}{T}\sum_{t \in \mathcal{T}}\mathbb{E}\big[Q^{\mathsf{rsu}}_k(t)\big]<\infty, \ \forall  k\in \mathcal{K}, \label{eq:Qrsu_constraint}\\[-.2em]
     & \lim_{T\rightarrow +\infty}\frac{1}{T}\sum_{t \in \mathcal{T}}\mathbb{E}\big[Q^{\mathsf{veh}}_j(t)\big]<\infty, \ \forall j\in \mathcal{J}\setminus \{1\}, \label{eq:Qveh_constraint}\\[-.2em] 
     &  \varphi_{i}(t)\in \{1,...,L_k+1\}, \  \forall t \in \mathcal{T}, i \in \mathcal{I}_k,\label{eq:partition_constraint}\\[-.2em]
     &  \xi_{i,j}(t)\in \{0,1\}, \ \forall t \in \mathcal{T}, i \in \mathcal{I}, j \in \mathcal{J},\label{eq:offload_constraint}\\[-.2em]
     &0 \leq f^{\mathsf{rsu}}_k(t) \leq f^{\mathsf{rsu,max}}, \ \forall t \in \mathcal{T}, k \in \mathcal{K}, \label{eq:singlefk_constraint}\\[-.2em]
     & \sum_{k \in \mathcal{K}}f^{\mathsf{rsu}}_k(t)\leq f^{\mathsf{rsu,max}}, \ \forall t \in \mathcal{T}, \label{eq:maxfk_constraint}
     \vspace{-1.5mm}
\end{align}
where the optimization variables are (i) $\bm{\varphi}(t)=\{\varphi_i(t)\}_{i\in\mathcal{I}}$ is the DNN partitioning decision vector for all CVs at time slot $t$; (ii) $\bm{\xi}(t)=\{\xi_{i,j}(t)\}_{i\in\mathcal{I},j\in\mathcal{J}}$ is a matrix describing the task offloading decisions for all CVs at time slot $t$; (iii) $\bm{F}(t)=\{f^{\mathsf{rsu}}_k(t)\}_{k\in\mathcal{K}}$ is the computation resource allocation vector for all DNN model types at time slot $t$. 

In $(\mathcal{P}_1)$, constraints~\eqref{eq:Qloc_constraint}-\eqref{eq:Qveh_constraint} guarantee the task queue stability of CVs, the RSU, and SVs\footnote{Expectation is taken with respect to the system random events~\cite{9449944}, e.g., vehicles' mobility and channel fading considered in this paper.}. Constraint~\eqref{eq:partition_constraint} implies that the DNN partitioning decision is integer. Constraint~\eqref{eq:offload_constraint} ensures that the task offloading decision is binary. Besides, constraint~\eqref{eq:singlefk_constraint} is the value range of the computation resources of the RSU allocated to different DNN model types, where $f^{\mathsf{rsu,max}}$ is the maximum computation capability of the RSU, and constraint~\eqref{eq:maxfk_constraint} guarantees adherence to the limited computation resources at the RSU.

\begin{remark}\label{rem:NPhard} Due to the non-linearity and recursive nature of long-term constraints~\eqref{eq:Qloc_constraint}-\eqref{eq:Qveh_constraint} -- as shown in~\eqref{eq:local_queue},~\eqref{eq:rsu_queue},~\eqref{eq:sv_queue} -- the DNN partitioning decisions, offloading decisions, and resource allocations among CVs are time coupled. Furthermore, due to the existence of both continues and discrete/binary variables as dictated by~\eqref{eq:partition_constraint},~\eqref{eq:offload_constraint}, and~\eqref{eq:singlefk_constraint}, problem $(\mathcal{P}_1)$ is a Mixed Integer NonLinear Program (MINLP), known to be NP-hard. Thus, it is hard to efficiently solve problem $(\mathcal{P}_1)$. 
\end{remark}
\vspace{-.5mm}

In the following, as the first step of our methodology, we develop a Lyapunov optimization technique to transform the long-term problem $(\mathcal{P}_1)$ into a per-slot deterministic problem, aiming at guaranteeing the long-term constrains. 

\vspace{-3mm}
\section{Lyapunov-Based Dynamic Long-Term Problem Decoupling}\label{sec:Lyapunov}
Henceforth, we first provide an overview of Lyapunov optimization related to our problem and then use the Lyapunov technique to decouple the formulated long-term MINLP problem $(\mathcal{P}_1)$ into deterministic problems on a per-slot basis.

\vspace{-3mm}
\subsection{Basics of Lyapunov Optimization}\label{subsec:basic_Lyapunov}
Lyapunov optimization~\cite{neely2022stochastic} is recognized as a powerful technique for decoupling a long-term stochastic optimization problem into sequential per-slot deterministic problems, while offering theoretical assurances for long-term system stability.

Tailoring this technique to our problem setup, Lyapunov optimization initially makes use of a Lyapunov function, which we obtain in~\eqref{eq:Lyapunovfunc}, to consolidate all the task queues. Following this, a Lyapunov drift function, which we obtain in~\eqref{eq:Lyapunovdrift}, is used to capture queue updates between two consecutive time slots. Subsequently, by minimizing the upper bound of the drift-plus-penalty expression, which we obtain in~\eqref{eq:upperbound}, for each time slot, we will be able to satisfy the dual goals of minimizing the long-term DNN-based task completion time while guaranteeing the system stability over time.

\vspace{-3mm}
\subsection{Transformation of Problem $(\mathcal{P}_1)$ via Lyapunov Technique}\label{subsec:transformation_Lyapunov}

First, we define a compact description of the backlog of task queues of CVs, the RSU, and SVs at time slot $t$ as follows:
\vspace{-1.5mm} \label{eq:total_queue}
\begin{align}
 \hspace{-1.5mm}   \bm{Q}(t)\!\!=\!\!\Big\{\! \{Q^{\mathsf{loc}}_i(t)\}_{i\in \mathcal{I}}, \{Q^{\mathsf{rsu}}_k(t)\}_{k\in \mathcal{K}}, \{Q^{\mathsf{veh}}_j(t)\}_{j\in \mathcal{J}\setminus \{1\}}\!\Big\}. 
    \vspace{-1.5mm}
\end{align}

Then, given that DNN-based tasks are generated at the beginning of each time slot, we use the quadratic form of the Lyapunov function, known for its reduced complexity in managing dynamic systems~\cite{10115012}, to collectively capture the status of all the task queues in $\bm{Q}(t)$, which is specified by
\vspace{-1.5mm}
\begin{align}\label{eq:Lyapunovfunc}
    \hspace{-0.5em} \bm{L}(\bm{Q}(t))\!=\! \frac{1}{2}\Bigg[\! \sum_{i\in \mathcal{I}}Q^{\mathsf{loc}}_i(t)^2 \!+\!\! \sum_{k \in \mathcal{K}}Q^{\mathsf{rsu}}_k(t)^2+\!\! \sum_{j\in \mathcal{J}\setminus \{1\}}\!\!\!\!Q^{\mathsf{veh}}_j(t)^2\! \Bigg].
    \vspace{-1.5mm}
\end{align}

Subsequently, we use the quadratic Lyapunov function in \eqref{eq:Lyapunovfunc} across two consecutive time slots to obtain the Lyapunov drift function that captures the updates in the states of the queues across consecutive time slots, which is given by
\vspace{-1.5mm}
\begin{equation}\label{eq:Lyapunovdrift}
    \Delta(\bm{Q}(t))= \mathbb{E}\Big[\bm{L}(\bm{Q}(t+1))-\bm{L}(\bm{Q}(t))\mid \bm{Q}(t) \Big].
    \vspace{-1.5mm}
\end{equation}
Here, a high value of the drift function implies a higher chance of instability in the task queues, and vice versa~\cite{neely2022stochastic}.

To combine the objective function of problem $(\mathcal{P}_1)$ with the stability of the system captured via~\eqref{eq:Lyapunovdrift}, we obtain the Lyapunov drift-plus-penalty function as
\vspace{-1.5mm}
\begin{align}\label{eq:Lyapunovdriftpluspenalty}
    \Lambda(\bm{Q}(t))=\Delta(\bm{Q}(t))+V\mathbb{E}\Big[\sum_{i \in \mathcal{I}}d_i(t)\mid\bm{Q}(t)\Big],
    \vspace{-1.5mm}
\end{align}
where $V$ is a tunable weight, representing the relative importance of task completion time compared to queue stability. We next obtain an upper-bound on~\eqref{eq:Lyapunovdriftpluspenalty}, which we will later use to transform the time-coupled problem $(\mathcal{P}_1)$ to a series of per time-slot problems.

\vspace{-1.5mm}
\begin{lemma}\label{lemma:upperbound}
For any feasible set of $\{\bm{\varphi}(t),\bm{\xi}(t),\bm{F}(t)\}_{t\in\mathcal{T}}$, which satisfies constraints~\eqref{eq:Qloc_constraint} to~\eqref{eq:maxfk_constraint}, the Lyapunov drift-plus-penalty function $\Lambda(\bm{Q}(t))$ can be upper bounded as follows:
\vspace{-1.5mm}
\begin{align}\label{eq:upperbound}
   &\Lambda(\bm{Q}(t))=\Delta(\bm{Q}(t))+V\mathbb{E}\Big[\sum_{i \in \mathcal{I}}d_i(t)\mid\bm{Q}(t)\Big]\nonumber\\
  &\leq \mathbb{E}\Bigg[\sum_{i\in \mathcal{I}}Q_i^{\mathsf{loc}}(t)\Big[\mathbb{I}_{\{\varphi_{i}(t)\neq1\}}\Big(\sum_{l_{k}=1}^{\varphi_{i}(t)-1}B_{l_{k}}\Big)-f^{\mathsf{loc}}_i\tau\Big]\nonumber\\
  &+\!\! \sum_{k \in\mathcal{K}}\!Q_k^{\mathsf{rsu}}(t)\!\Big[\!\sum_{i \in \mathcal{I}_k}\xi_{i,1}(t)\mathbb{I}_{\{\varphi_{i}(t)\neq L_{k}+1\}}\!\Big(\!\sum_{l_{k}=\varphi_{i}(t)}^{L_{k}}B_{l_{k}}\!\Big)\!\!-\!f^{\mathsf{rsu}}_k(t)\tau\Big]  \nonumber\\
  &+\!\!\!\!\!\!\sum_{j\in \mathcal{J}\setminus\{1\}}\!\!\!\!\!Q_j^{\mathsf{veh}}(t)\!\Big[\sum_{k \in \mathcal{K}}\sum_{i \in \mathcal{I}_k}\!\xi_{i,j}(t)\mathbb{I}_{\{\varphi_{i}(t)\neq L_{k}+1\}}\!\Big(\!\sum_{l_{k}=\varphi_{i}(t)}^{L_{k}}\!\!\!\!\!B_{l_{k}}\!\Big)\!\!-\!f^{\mathsf{veh}}_j\tau\!\Big]\nonumber\\
  &+V\sum_{i \in \mathcal{I}}d_i(t)\Big| \bm{Q}(t)\Bigg]+\chi,
  \vspace{-.5mm}
\end{align}
where $d_i(t)$ is given in~\eqref{eq:total_delay}, and $\chi$ is a constant given in~\eqref{eq:constant_value}, which is shown at the top of the previous page.

\end{lemma}
\vspace{-2.3mm}
\begin{proof} 
The proof is provided in Appendix~\ref{app:upperbound}
\end{proof}

By omitting the constant component $\chi$, which is not impacted by the task queues, we transform the original problem $(\mathcal{P}_1)$ into the subsequent per-slot deterministic optimization problem $(\mathcal{P}_2(t))$ which can be solved at each time slot without the need of knowing the future realizations of random channel conditions and vehicles' mobility, while satisfying the long-term constraints and ensuring the stable system operation:
\vspace{-1.5mm}
\begin{align}
    &(\mathcal{P}_2(t)): \min_{\{\bm{\varphi}(t),\bm{\xi}(t),\bm{F}(t)\}}V\sum_{i \in \mathcal{I}}d_i(t)\nonumber\\
    &+\sum_{i\in \mathcal{I}}Q_i^{\mathsf{loc}}(t)\Big[\mathbb{I}_{\{\varphi_{i}(t)\neq1\}}\Big(\sum_{l_{k}=1}^{\varphi_{i}(t)-1}B_{l_{k}}\Big)-f^{\mathsf{loc}}_i\tau\Big]\nonumber\\
  &+\!\! \sum_{k \in\mathcal{K}}\!Q_k^{\mathsf{rsu}}(t)\!\Big[\!\sum_{i \in \mathcal{I}_k}\xi_{i,1}(t)\mathbb{I}_{\{\varphi_{i}(t)\neq L_{k}+1\}}\!\Big(\!\sum_{l_{k}=\varphi_{i}(t)}^{L_{k}}B_{l_{k}}\!\Big)\!\!-\!f^{\mathsf{rsu}}_k(t)\tau\Big]\nonumber\\
  &+\!\!\!\! \sum_{j\in \mathcal{J}\setminus\{1\}}\!\!\!\!\!\!Q_j^{\mathsf{veh}}(t)\!\Big[\sum_{k \in \mathcal{K}}\sum_{i \in \mathcal{I}_k}\!\xi_{i,j}(t)\mathbb{I}_{\{\varphi_{i}(t)\neq L_{k}+1\}}\!\Big(\!\sum_{l_{k}=\varphi_{i}(t)}^{L_{k}}\!\!\!\!B_{l_{k}}\!\Big)\!\!-\!\!f^{\mathsf{veh}}_j\tau\!\Big]\label{eq:problem_2} \\[-.15em]
     & \textrm{s.t.}~~\text{Constraints}~\eqref{eq:partition_constraint}-\eqref{eq:maxfk_constraint}~\text{in}~(\mathcal{P}_1), \nonumber
\end{align}
where $d_i(t)$ is given in~\eqref{eq:total_delay}. Our subsequent aim is thus to solve $(\mathcal{P}_2(t))$ for each time slot. Note that problem $(\mathcal{P}_2(t))$ is a mixed-integer programming problem because: (i) $\{\bm{\varphi}(t),\bm{\xi}(t),\bm{F}(t)\}$ are a mixture of discrete and continuous variables; (ii) the objective function is non-convex. Existing approaches, which include heuristic-based~\cite{8854339},~\cite{8451923} and decomposition-oriented search algorithms~\cite{8533343}, are either time consuming to solve $(\mathcal{P}_2(t))$ due to their high complexity or lead to a relatively weak solution especially as the problem size grows. To this end, in the following, we introduce a Multi-Agent Diffusion-based Deep Reinforcement Learning (MAD2RL) algorithm to tackle $(\mathcal{P}_2(t))$.

\vspace{-3mm}
\section{Basic Idea of Diffusion Models}\label{sec:DDPM}
\vspace{-.1mm}
Henceforth, before delving into the Multi-Agent Diffusion-based Deep Reinforcement Learning (MAD2RL) algorithm, we first give our motivation of adopting the diffusion model. We then elaborate on the diffusion model customization for generating optimal decisions regarding DNN partitioning and task offloading. Note that a closed-form for resource allocation will be obtained through convex optimization techniques once the above decisions are known (details will be provided in Sec.~\ref{subsec:subroutine}).

\vspace{-3mm}
\subsection{Motivation of Adopting Diffusion Model}  \label{subsec:DDPM_motivation}  
Recently, DRL has been recognized as a promising approach to tackle a variety of networking problems in dynamic computing environments~\cite{9234039,10056271,9552188},~\cite{10497174}. Generally speaking, DRL utilizes DNNs to learn the optimal action given the state of the system. The Multi-Layer Perceptron (MLP), a type of fully connected DNN that comprises various hidden layers with nonlinear activation functions, has been extensively employed in the DRL architecture. However, in this paper, problem $(\mathcal{P}_2(t))$ presents several unique challenges: (i) the inherent mobility of vehicles introduces uncertainties into DRL environments, making the state space complex and highly dynamic across different time slots; (ii) the discrete variables, specifically $\{\bm{\varphi}(t), \bm{\xi}(t)\}$, are intertwined. This combinatorial nature of variables can cause the solution space to expand exponentially as the number of CVs increases. Consequently, the performance of MLP may diminish in environments characterized by complex and high-dimensional state and action spaces~\cite{8103164}.

Diffusion models present several compelling advantages over MLPs, especially valuable in uncertain and complex DRL environments such as those encountered in VEC. Specifically, the generative capabilities of diffusion models not only enhance the action sample efficiency by gradually removing noise across various denoising steps (to be introduced later), but also provide a richer understanding of the environmental state through their superior feature representation capabilities. As a result, the use of diffusion models is significantly beneficial for solving the complex problem of $(\mathcal{P}_2(t))$.

\vspace{-3mm}
\subsection{Preliminaries of Diffusion Model}  \label{subsec:DDPM} 

Drawing inspiration from the Denoising Diffusion Probabilistic Model (DDPM)~\cite{ho2020denoising}, we aim to design a framework for generating optimal decisions regarding DNN partitioning and task offloading. Specifically, DDPM encompasses two pivotal processes: the \emph{forward process}, which gradually introduces noise to the optimization solution at each noising step until it becomes indistinguishable from Gaussian noise, and the \emph{reverse process}, which gradually removes noise at each denoising step to reconstruct the optimization solution from its noisy counterpart. 

In this study, the optimal DNN partitioning decision $\varphi^*_{i}(t)\in \{1,...,L_k+1\}$ and task offloading decision $\xi^*_{i}(t)\in\{1,...,J\}$\footnote{We make a slight adjustment to the DNN task offloading decision in our simulation. Here, $\forall i \in \mathcal{I},\ \xi_{i}(t)\in\{1,...,J\}$, where $\xi_{i}(t)=j$ signifies that $\xi_{i,j}(t)=1$ within our proposed system model.} of CV $i\in\mathcal{I}_k$ at time slot $t$ are integrated into $(L_k+1)J$ elements. The probability distribution $\bm{x}_i^0(t)\sim\mathbb{R}^{(L_k+1)J}$ of each decision being selected is regarded as the optimization solution in the context of DDPM.
Then, according to the DDPM, $\bm{x}_i^0(t)$ will be gradually added with noise at each noising step until it becomes Gaussian noise, known as the forward process. Subsequently, in the reverse process, the optimal decision generation network $\mathcal{Q}_{\bm{\theta}_i}(\cdot)$ is viewed as a denoiser which starts from Gaussian noise and gradually recovers $\bm{x}_i^0(t)$ with local state $\bm{s}_i(t)$ (will be formulated later in Sec. \ref{subsec:MDP}) as the input. 

\begin{figure}[t]
\vspace{-1mm}
\includegraphics[width=.48\textwidth]{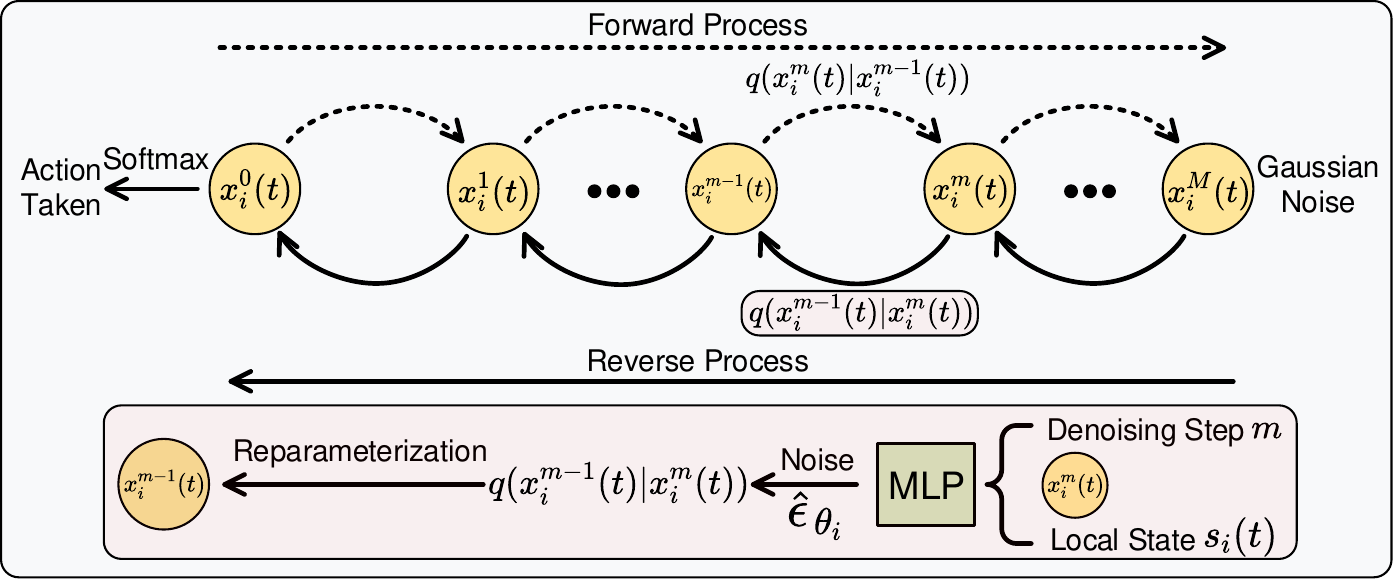}
\centering
\vspace{-1.5mm}
\caption{Illustration of the diffusion model tailored for generating the optimal decisions of DNN partitioning and task offloading for CV $i$ at time slot $t$.}
\label{fig:DDPM}
\end{figure}
\vspace{-2mm}
\subsubsection{Forward Process}  \label{subsubsec:forward_process}  
A schematic of our method using diffusion model tailored for generating the optimal decisions of DNN partitioning and task offloading for CV $i$ at time slot $t$ is depicted in Fig.~\ref{fig:DDPM}. Let $\mathcal{M}=\{1,...,M\}$ denote the set of noising/denoising steps, at each time slot $t$, for CV $i$, given the probability distribution  $\bm{x}^0_i(t)$, the forward process adds a sequence of Gaussian noises at each noising step to obtain $\bm{x}^1_i(t),...,\bm{x}^M_i(t)$. The transition from $\bm{x}^{m-1}_i(t)$ to $\bm{x}^{m}_i(t)$ is defined as a normal distribution with mean $\sqrt{1-\beta_m}\bm{x}^{m-1}_i(t)$ and variance $\beta_m\mathbf{I}$ given by~\cite{ho2020denoising}
\vspace{-1.5mm}
\begin{align}\label{eq:forward_distribution}
    &q(\bm{x}^m_i(t)|\bm{x}^{m-1}_i(t))\nonumber\\
    &= \mathcal{N}(\bm{x}^m_i(t);\sqrt{1-\beta_m}\bm{x}^{m-1}_i(t),\beta_m\mathbf{I}),
    \vspace{-1.5mm}
\end{align}
where $\beta_m$ is the diffusion rate at noising step $m$~\cite{ho2020denoising} calculated by $\beta_m=1- e^{-\frac{\beta^{\mathsf{min}}}{M}-\frac{2m-1}{2M^2}(\beta^{\mathsf{max}}-\beta^{\mathsf{min}})}$, with $\beta^{\mathsf{min}}$ and $\beta^{\mathsf{max}}$ are the predetermined minimum and maximum diffusion rates, respectively, and $\mathbf{I}$ is an identity matrix.

Since $\bm{x}^m_i(t)$ is sampled from the normal distribution $\mathcal{N}(\sqrt{1-\beta_m}\bm{x}^{m-1}_i(t),\beta_m\mathbf{I})$, i.e., $\mathcal{N}(\bm{x}^m_i(t);\sqrt{1-\beta_m}\bm{x}^{m-1}_i(t),\beta_m\mathbf{I}) \equiv \bm{x}^m_i(t)\sim \mathcal{N}(\sqrt{1-\beta_m}\bm{x}^{m-1}_i(t),\beta_m\mathbf{I})$, we can obtain the mathematical relationship between $\bm{x}^{m-1}_i(t)$ and $\bm{x}^m_i(t)$ via reparameterization technique as follows~\cite{ho2020denoising}:
\vspace{-1.5mm}
\begin{equation}\label{eq:forward_update}
    \bm{x}^m_i(t)= \sqrt{1-\beta_m}\bm{x}^{m-1}_i(t)+\sqrt{\beta_m}\bm{\epsilon}_{m-1},
    \vspace{-1.5mm}
\end{equation}
where $\bm{\epsilon}_{m-1}$ is the Gaussian noise sampled from the standard normal distribution $\mathcal{N}(0,\mathbf{I})$. Finally, based on~\eqref{eq:forward_update}, the mathematical relationship between $\bm{x}^0_i(t)$ and $\bm{x}^m_i(t)$ at any noising step $m$ can be calculated as
\vspace{-1.5mm}
\begin{align}\label{eq:forward_relationship}
    \bm{x}^m_i(t)=\sqrt{\hat{\alpha}_m}\bm{x}^{0}_i(t)+\sqrt{1-\hat{\alpha}_m}\bm{\epsilon}_m,
    \vspace{-1.5mm}
\end{align}
where $\hat{\alpha}_m=\prod_{u=1}^{m}\alpha_u$ is the cumulative product of $\alpha_u$ over previous nosing steps $m$, $\alpha_u=1-\beta_u$, and $\bm{\epsilon}_m \sim \mathcal{N}(0,\mathbf{I})$. As observed from~\eqref{eq:forward_relationship}, $\bm{x}^M_i(t)$ transitions to pure noise following a normal distribution $\mathcal{N}(0,\mathbf{I})$, starting from the initial probability distribution $\bm{x}^0_i(t)$ as the noising step increases.

In this paper, the diffusion model is leveraged to generate the probability distribution $\bm{x}^0_i(t)$ without having it as the optimization solution a priori. Therefore, the forward process serves solely to establish the relationship between $\bm{x}^0_i(t)$ and $\bm{x}^m_i(t)$ as presented in~\eqref{eq:forward_relationship}, which is essential for the reverse process described below. As such, the forward process is not actually performed in this work and thus is depicted with a dotted line in Fig.~\ref{fig:DDPM}.
\vspace{-2mm}
\subsubsection{Reverse Process}  \label{subsubsec:reverse_process} 
At each time slot $t$, for CV $i$, the reverse process aims to infer the probability distribution $\bm{x}_i^0(t)\sim\mathbb{R}^{(L_k+1)J}$ of each decision being selected from a noise sample $\bm{x}^M_i(t)\sim\mathcal{N}(0,\mathbf{I})$. Specifically, according to~\eqref{eq:forward_relationship}, we reconstruct the relationship between $\bm{x}^0_i(t)$ and $\bm{x}^m_i(t)$ as
\vspace{-1.5mm}
\begin{align}\label{eq:reverse_relationship}
    \bm{x}^0_i(t)= \frac{1}{\sqrt{\hat{\alpha}_m}}(\bm{x}^m_i(t)-\sqrt{1-\hat{\alpha}_m}\bm{\hat{\epsilon}}_m).
    \vspace{-1.5mm}
\end{align}
Here, $\bm{\hat{\epsilon}}_m$ represents a new source of noise at each denoising step $m$, leveraged to reconstruct the probability distribution $\bm{x}_i^0(t)$, which is independent of the noise $\bm{\epsilon}_m$ introduced in the forward process. 

As a result, $\bm{\hat{\epsilon}}_m$ can be learned by a deep neural network using $\bm{x}^{m}_i(t)$, denoising step $m$, and local state $\bm{s}_i(t)$ as inputs. In DDPM, the optimization objective is the Mean Squared Error (MSE) loss between the noise $\bm{\epsilon}_m$ introduced during the forward process and the noise $\bm{\hat{\epsilon}}_m$ learned by a deep model at each step. However, since the forward process is not actually conducted in this paper, the training objective of the reverse process will experience a shift towards minimizing the objective function given in~\eqref{eq:problem_2} in an exploratory manner (as detailed in Sec.~\ref{subsec:D2RL_architecture}).

To infer the probability distribution $\bm{x}_i^0(t)$ through various denoising steps, we establish the transition from $\bm{x}^{m}_i(t)$ to $\bm{x}^{m-1}_i(t)$, which has been proven to follow a Gaussian distribution as follows~\cite{ho2020denoising}:
\vspace{-1.5mm}
\begin{align}\label{eq:reverse_distribution}
q(\bm{x}^{m-1}_i(t)|\bm{x}^m_i(t))= \mathcal{N}(\bm{x}^{m-1}_i(t);\bm{\mu}^m_i(t),\hat{\beta}_m\mathbf{I}),
\vspace{-1.5mm}
\end{align}
where $\hat{\beta}_m=\frac{1-\hat{\alpha}_{m-1}}{1-\hat{\alpha}_m}\beta_m$, and mean $\bm{\mu}^m_i(t)$ can be obtained through the Bayesian formula as follows~\cite{ho2020denoising}:
\vspace{-1.5mm}
\begin{align}\label{eq:reverse_original_mean}
\bm{\mu}^m_i(t)= \frac{\sqrt{\alpha_m}(1-\hat{\alpha}_{m-1})}{1-\hat{\alpha}_m}\bm{x}^{m}_i(t)+\frac{\sqrt{\hat{\alpha}_{m-1}}\beta_m}{1-\hat{\alpha}_m}\bm{x}^{0}_i(t).
\vspace{-1.5mm}
\end{align}

Subsequently, incorporating~\eqref{eq:reverse_relationship} into~\eqref{eq:reverse_original_mean} and utilizing a deep model to learn $\bm{\hat{\epsilon}}_m$, the mean $\bm{\mu}^m_i(t)$ can be obtained as follows (hereafter we change the form of $\bm{\mu}^m_i(t)$ accordingly)
\vspace{-1.5mm}
\begin{align}\label{eq:reverse_mean}
&\bm{\mu}^m_{\bm{\theta}_i}(\bm{x}^{m}_i(t),m,\bm{s}_i(t))\nonumber\\
&=\frac{1}{\sqrt{\alpha_m}}\Big[\bm{x}^{m}_i(t)-\frac{1-\alpha_m}{\sqrt{1-\hat{\alpha}_m}}\bm{\hat{\epsilon}}_{\bm{\theta}_i}(\bm{x}^{m}_i(t),m,\bm{s}_i(t))\Big].
\vspace{-1.5mm}
\end{align}
Then, based on~\eqref{eq:reverse_distribution}, by employing the reparameterization technique, the mathematical relationship between $\bm{x}^m_i(t)$ and $\bm{x}^{m-1}_i(t)$ can be obtained by
\vspace{-1.5mm}
\begin{equation}\label{eq:reverse_update}
    \bm{x}^{m-1}_i(t)= \bm{\mu}^m_{\bm{\theta}_i}(\bm{x}^{m}_i(t),m,\bm{s}_i(t))+\sqrt{\hat{\beta}_m}\bm{\epsilon}_m,
    \vspace{-1.5mm}
\end{equation}
where $\bm{\epsilon}_m \sim \mathcal{N}(0,\mathbf{I})$. By iteratively applying the reverse update rule in~\eqref{eq:reverse_update}, we can obtain the probability distribution $\bm{x}^0_m(t)$ after $M$ denoising steps.

Finally, the softmax function is leveraged to convert $\bm{x}^0_m(t)$ into a probability distribution as
\vspace{-1.5mm}
\begin{align}\label{eq:softmax_distribution}
    &\mathcal{Q}_{\bm{\theta}_i}(\bm{s}_i(t))\nonumber\\
    &= \Bigg\{ \frac{e^{\bm{x}^{0,u}_i(t)}}{\sum_{v=1}^{(L_k+1)J}e^{\bm{x}^{0,v}_i(t)}}, \forall u \in\{1,...,(L_k+1)J\} \Bigg\},
    \vspace{-1.5mm}
\end{align}
where the elements in $\mathcal{Q}_{\bm{\theta}_i}(\bm{s}_i(t))$ indicate the corresponding probability of selecting each action.
\vspace{-3mm}
\section{Multi-Agent Diffusion-Based Deep Reinforcement Learning (MAD2RL) Algorithm}\label{sec:D2RL}
\vspace{-.1mm}
Henceforth, we first provide an overview of the proposed MAD2RL algorithm followed by the motivation of adopting the MAD2RL algorithm based on the QMIX framework~\cite{rashid2020monotonic}. Subsequently, we propose an optimization subroutine for resource allocation that is based on convex optimization techniques. We next model the problem as a Markov Decision Process (MDP). We then present the architecture of our MAD2RL algorithm. Finally, we provide an analysis of the computational complexity associated with the proposed method.

\vspace{-3mm}
\subsection{Overview of The Proposed MAD2RL Algorithm} \label{subsec:d2rl_overview}
Generally speaking, our MAD2RL algorithm is based on the QMIX~\cite{rashid2020monotonic} framework, one of the state-of-the-art Multi-Agent Deep Reinforcement Learning (MADRL) schemes. In this framework, each CV is considered an agent that makes local action decisions based on its local observation. However, rather than directly optimizing discrete and continuous variables simultaneously, we further enhance the MAD2RL by incorporating a convex optimization technique as a subroutine, thereby improving its learning efficiency. This approach involves decoupling the optimization variables into two categories, which are then solved separately.

Specifically, we decompose the optimization variables $\{\bm{\varphi}(t),\bm{\xi}(t),\bm{F}(t)\}$ of problem $(\mathcal{P}_2(t))$ into two categories: $\{\bm{\varphi}(t),\bm{\xi}(t)\}$ and $\bm{F}(t)$. The variables $\{\bm{\varphi}(t),\bm{\xi}(t)\}$ are subsequently determined through an exploratory process using the diffusion model (to be discussed in Sec.\ref{subsec:D2RL_architecture}). With $\{\bm{\varphi}(t),\bm{\xi}(t)\}$ known, we find that determining of $\bm{F}(t)$ entails solving a convex optimization problem, for which a closed-form solution can be derived (details will be provided in Sec.\ref{subsec:subroutine}).

\vspace{-3mm}
\subsection{Motivation of Adopting QMIX Framework}  \label{subsec:DDPG_motivation} 
Recently, numerous edge computing and network optimization solutions have been proposed~\cite{10497174},~\cite{10472663, 10032267}, utilizing single agent DRL techniques such as Double Deep Q-Network (DDQN), DDPG, and Proximal Policy Optimization (PPO). However, these methods are challenging to apply in this study due to several reasons: (i) a single agent often struggles to show a satisfactory performance within environments with large state-action spaces. If single-agent DRL techniques are employed, the action space is of $((L_k+1)J)^I$ dimensions that expands exponentially as the number of CVs increases; (ii) a single agent will need to have the perfect information about the environment of all CVs (e.g., their channel conditions and task queues) to make an action, obtaining of which will impose notable overheads. 

As a result, we propose our MAD2RL algorithm, which is based on the QMIX~\cite{rashid2020monotonic} framework. Benefiting from MADRL techniques, MAD2RL algorithm not only effectively addresses the issue of action space explosion by decomposing the multiple objectives among different CVs (agents), but also facilitates the cooperative learning of DNN partitioning, task offloading, and resource allocation policy across different CVs (agents).

\vspace{-3mm}
\subsection{Optimization Subroutine for Resource Allocation}  \label{subsec:subroutine}  
After a close observation of problem $(\mathcal{P}_2(t))$, we notice that only the edge processing delay at the RSU, i.e., $d^{\mathsf{rsu}}_i(t)$ and the task queue at the RSU, i.e., $Q^{\mathsf{rsu}}_k(t)$ are impacted by the computing resource allocation. Additionally, the aggregated delay from the perspective of all CVs is equivalent to the aggregated delay from the perspective of all DNN model types, i.e., $\sum_{i \in \mathcal{I}}d_i(t) \equiv \sum_{k \in \mathcal{K}}\sum_{i \in \mathcal{I}_k}d_i(t)$. Then, given DNN partitioning decisions $\bm{\varphi}^*(t)$ and task offloading decisions $\bm{\xi}^*(t)$, which will be later obtained by our MAD2RL method, the optimal computation resource allocation problem $(\mathcal{P}_3(t))$ can be deducted from $(\mathcal{P}_2(t))$ as follows:
\vspace{-1.5mm}
\begin{align}
     (\mathcal{P}_3(t)):&\min_{\bm{F}(t)}   V\sum_{k\in\mathcal{K}}\sum_{i\in\mathcal{I}^*_{k}(t)}\Bigg[\frac{Q^{\mathsf{rsu}}_k(t)+\sum_{l_k=\varphi^*_{i}(t)}^{L_{k}}B_{l_k}
  }{f^{\mathsf{rsu}}_k(t)}  \nonumber\\
  &+\frac{\sum_{i^{\prime}\in \mathcal{I}^*_{k}(t)\setminus \{i\}}\Big(\sum_{l_{k}=\varphi^*_{i^{\prime}}(t)}^{L_{k}}B_{l_{k}}\Big)}{2f^{\mathsf{rsu}}_k(t)}\Bigg] \nonumber\\
  &-\sum_{k \in \mathcal{K}}Q^{\mathsf{rsu}}_k(t)f^{\mathsf{rsu}}_k(t)\tau   \label{eq:problem_3}\\ 
     &\textrm{s.t.} \nonumber\\
     &0 \leq f^{\mathsf{rsu}}_k(t) \leq f^{\mathsf{rsu,max}}, \ \forall t \in \mathcal{T}, k \in \mathcal{K}, \label{eq:p3_st1}\\[-.2em]
     & \sum_{k\in\mathcal{K}}f^{\mathsf{rsu}}_k(t)\leq f^{\mathsf{rsu,max}}, \ \forall t \in \mathcal{T}, k \in \mathcal{K}, \label{eq:p3_st2}
\end{align}
where $\mathcal{I}^*_{k}(t)$ is the set of CVs $i$ with the $k$th type DNN model who offload their tasks to the RSU at time slot $t$ with $\varphi_i(t)\neq L_k+1$. Notice that the constraints in~\eqref{eq:p3_st1} and~\eqref{eq:p3_st2} are linear. Denoting the objective function in~\eqref{eq:problem_3} as $\Psi(\bm{F}(t))$, by calculating its corresponding second-order derivatives w.r.t. $f^{\mathsf{rsu}}_k(t)$, we can obtain
\vspace{-1.5mm}
\begin{equation}\label{eq:secondorder_p3}
    \frac{\partial^2\Psi(\bm{F}(t))}{\partial f^{\mathsf{rsu}}_k(t)^2}=\frac{2\Gamma_k(t)}{f^{\mathsf{rsu}}_k(t)^3}>0,
    \vspace{-.5mm}
\end{equation}
where 
\vspace{-2.5mm}
\begin{align}
    \Gamma_k(t)= VQ^{\mathsf{rsu}}_k(t)+V\sum_{i\in\mathcal{I}^*_{k}(t)}\Big(\sum_{l_k=\varphi^*_{i}(t)}^{L_{k}}\!\!B_{l_k}\Big) \nonumber \\
    \vspace{-1.5em}
    +\frac{V}{2}\sum_{i\in\mathcal{I}^*_{k}(t)}\sum_{i^{\prime}\in \mathcal{I}^*_{k}(t)\setminus \{i\}}\!\!\Big(\sum_{l_{k}=\varphi^*_{i^{\prime}}(t)}^{L_{k}}\!\!B_{l_{k}}\Big).
\vspace{-.5mm}
\end{align}
Thus, $(\mathcal{P}_3(t))$ is a convex optimization problem and can be efficiently solved by Karush-Kuhn-Tucker (KKT) conditions~\cite{8533343} as below.

First, taking constraint~\eqref{eq:p3_st2} into account, the Lagrangian function of problem $(\mathcal{P}_3(t))$ can be calculated by\footnote{Here, we slightly abuse index $\mathcal{L}$, initially leveraged to represent the set of DNN layers, to denote the Lagrangian.}
\vspace{-1.5mm}
\begin{flalign}
\mathcal{L}(\Psi(\bm{F}(t)),\eta(t))&= \sum_{k\in\mathcal{K}}\Big(\frac{\Gamma_k(t)}{f^{\mathsf{rsu}}_k(t)}-Q^{\mathsf{rsu}}_k(t)f^{\mathsf{rsu}}_k(t)\tau\Big)\nonumber\\
&+\eta(t)(\sum_{k \in \mathcal{K}}f^{\mathsf{rsu}}_k(t)-f^{\mathsf{rsu,max}}),	
\vspace{-.5mm}
\end{flalign}
where $\eta(t)$ is the Lagrangian multiplier at time slot $t$. Taking the derivatives of the Lagrangian w.r.t. $f^{\mathsf{rsu}}_k(t)$, we obtain 
\vspace{-1.5mm}
\begin{flalign} \label{eq:optalpha}
	\frac{\partial\mathcal{L}(\Psi(\bm{F}(t)),\eta(t))}{\partial f^{\mathsf{rsu}}_k(t)}=-\frac{\Gamma_k(t)}{f^{\mathsf{rsu}}_k(t)^2}-Q^{\mathsf{rsu}}_k(t)\tau+\eta(t).
\vspace{-.5mm}
\end{flalign}
Then, by equating~\eqref{eq:optalpha} to be zero and solving
for $f^{\mathsf{rsu}}_{k}(t)$, the optimal computation resource allocation can be obtained by
\vspace{-1.5mm}
\begin{flalign} \label{eq:nu}
	f^{\mathsf{rsu},*}_{k}(t)= \sqrt{\frac{\Gamma_k(t)}{\eta^*(t)-Q^{\mathsf{rsu}}_k(t)\tau}},
\vspace{-.5mm}
\end{flalign}
where $\eta^*(t)>Q^{\mathsf{rsu}}_k(t)\tau$ is a constant satisfying
\vspace{-1.5mm}
\begin{flalign} \label{eq:nu_constraint}
	\eta^*(t)(\sum_{k \in \mathcal{K}}f^{\mathsf{rsu},*}_k(t)-f^{\mathsf{rsu,max}})=0.
\vspace{-.5mm}
\end{flalign}
Subsequently, by substituting~\eqref{eq:nu} into~\eqref{eq:nu_constraint}, we have
\vspace{-1.5mm}
\begin{flalign} \label{eq:optnu}
    \sum_{k\in\mathcal{K}}\sqrt{\frac{\Gamma_k(t)}{\eta^*(t)-Q^{\mathsf{rsu}}_k(t)\tau}}=f^{\mathsf{rsu,max}}.
\vspace{-.5mm}
\end{flalign}
Consequently, the optimal Lagrangian multiplier $\eta^*(t)$ can be determined with low computational complexity using the bisection method~\cite{8533343}, considering~\eqref{eq:nu_constraint} and~\eqref{eq:optnu}, and the closed-form of the computation resource allocation $f^{\mathsf{rsu},*}_k(t)$ can be deduced from~\eqref{eq:nu}.

Therefore, when $\{\bm{\varphi}^*(t), \bm{\xi}^*(t)\}$ are efficiently solved by the MAD2RL (will be introduced later), the aforementioned subroutine can significantly decrease the associated training complexity by eliminating the need to search for $f_k^{\mathsf{rsu}}(t)$ within the solution space of the MAD2RL.

\vspace{-3mm}
\subsection{Formulation of MDP}  \label{subsec:MDP}
We first formulate the sequential decision making process in our problem of interest as a Markov Decision Process (MDP), which includes the \emph{state space}, \emph{action space}, and \emph{reward function}. Each element of the MDP is described next.
\vspace{-3mm}
\subsubsection{State Space} \label{subsubsec:state_space} Each CV is controlled by a dedicated agent. At each time slot $t$, the local state $\bm{s}_i(t)$ of agent $i$ includes the status of the local task queue of CV $i$: $Q^{\mathsf{loc}}_i(t)$, the task queue of the RSU regarding current DNN model type $i\in \mathcal{I}_k$: $Q^{\mathsf{rsu}}_k(t)$, the task queue of each SV: $\bm{Q}^{\mathsf{veh}}(t)=\{Q^{\mathsf{veh}}_j(t)\}_{j\in \mathcal{J}\setminus\{1\}}$, the instantaneous location of CV $i$: $po^{\mathsf{loc}}_i(t)$, and the instantaneous location of each SV: $\bm{P}^{\mathsf{veh}}(t)=\{po^{\mathsf{veh}}_j(t)\}_{j\in \mathcal{J}\setminus\{1\}}$. Therefore, the local state $\bm{s}_i(t)$ contains $2J+1$ elements, and is defined as follows:
\vspace{-1.5mm}
\begin{align}\label{eq:MDP_state}
    \bm{s}_i(t)=\{Q^{\mathsf{loc}}_i(t),Q^{\mathsf{rsu}}_k(t),\bm{Q}^{\mathsf{veh}}(t),po^{\mathsf{loc}}_i(t),\bm{P}^{\mathsf{veh}}(t)\}.
    \vspace{-.5mm}
\end{align}
Then, the joint state of all agents is defined as below:
\vspace{-1.5mm}
\begin{align}\label{eq:MDP_joint_state}
    \bm{S}(t)=\{\bm{s}_1(t),...,\bm{s}_i(t),...,\bm{s}_I(t)\}.
    \vspace{-.5mm}
\end{align}
\vspace{-3mm}
\subsubsection{Action Space} After obtaining local state $\bm{s}_i(t)$, agent $i \in \mathcal{I}_k$ will take its local action $a_i(t)$ according to the probability distribution $\bm{x}_i^0(t)\sim\mathbb{R}^{(L_k+1)J}$, which can be given by
\vspace{-1.5mm}
\begin{align}\label{eq:MDP_action}
    a_i(t) \!=\! \argmax_u \!\big\{\mathcal{Q}^u_{\bm{\theta}_i}(\bm{s}_i(t)), \forall u \in\{1,...,(L_k+1)J\} \!\big\},
    \vspace{-.5mm}
\end{align}
where $\varphi_{i}(t) = \lfloor \frac{a_i(t)}{J} \rfloor$, and $\xi_{i}(t) =  a_i(t) \mod J$. Then, joint action of all agents is defined as below:
\vspace{-1.5mm}
\begin{align}\label{eq:MDP_joint_action}
    \bm{A}(t)=\{a_1(t),...,a_i(t),...,a_I(t)\}.
    \vspace{-.5mm}
\end{align}
\vspace{-3mm}
\subsubsection{Reward Function} After agent $i$ executes action $a_i(t)$ at state $\bm{s}_i(t)$, the environment transitions to a new state $\bm{s}_i(t+1)$ and provides a feedback in the form of a reward. Given that our problem of interest is situated within a fully cooperative multi-agent scenario~\cite{rashid2020monotonic}, a group of agents collaboratively work towards optimizing a common reward. Consequently, as the goal of this paper is to minimize the objective function presented in~\eqref{eq:problem_2}, we obtain the explicit formulation of the common reward function $r(t)$ for each agent $i$ as
\vspace{-1.5mm}
\begin{align}\label{eq:MDP_reward}
    &r(t)= -\sum_{i\in \mathcal{I}}Q_i^{\mathsf{loc}}(t)\Big[\mathbb{I}_{\{\varphi_{i}(t)\neq1\}}\Big(\sum_{l_{k}=1}^{\varphi_{i}(t)-1}B_{l_{k}}\Big)-f^{\mathsf{loc}}_i\tau\Big]\nonumber\\
  &-\!\! \sum_{k \in\mathcal{K}}\!Q_k^{\mathsf{rsu}}(t)\!\Big[\!\sum_{i \in \mathcal{I}_k}\xi_{i,1}(t)\mathbb{I}_{\{\varphi_{i}(t)\neq L_{k}+1\}}\!\Big(\!\sum_{l_{k}=\varphi_{i}(t)}^{L_{k}}B_{l_{k}}\!\Big)\!\!-\!f^{\mathsf{rsu}}_k(t)\tau\Big]\nonumber\\
  &-\!\!\!\! \sum_{j\in \mathcal{J}\setminus\{1\}}\!\!\!\!\!\!Q_j^{\mathsf{veh}}(t)\!\Big[\sum_{k \in \mathcal{K}}\sum_{i \in \mathcal{I}_k}\!\xi_{i,j}(t)\mathbb{I}_{\{\varphi_{i}(t)\neq L_{k}+1\}}\!\Big(\!\sum_{l_{k}=\varphi_{i}(t)}^{L_{k}}\!\!\!\!B_{l_{k}}\!\Big)\!\!-\!\!f^{\mathsf{veh}}_j\tau\!\Big]\nonumber\\
  &-V\sum_{i \in \mathcal{I}}d_i(t).
\end{align}

\begin{figure}[t]
\vspace{-1mm}
\includegraphics[width=.48\textwidth]{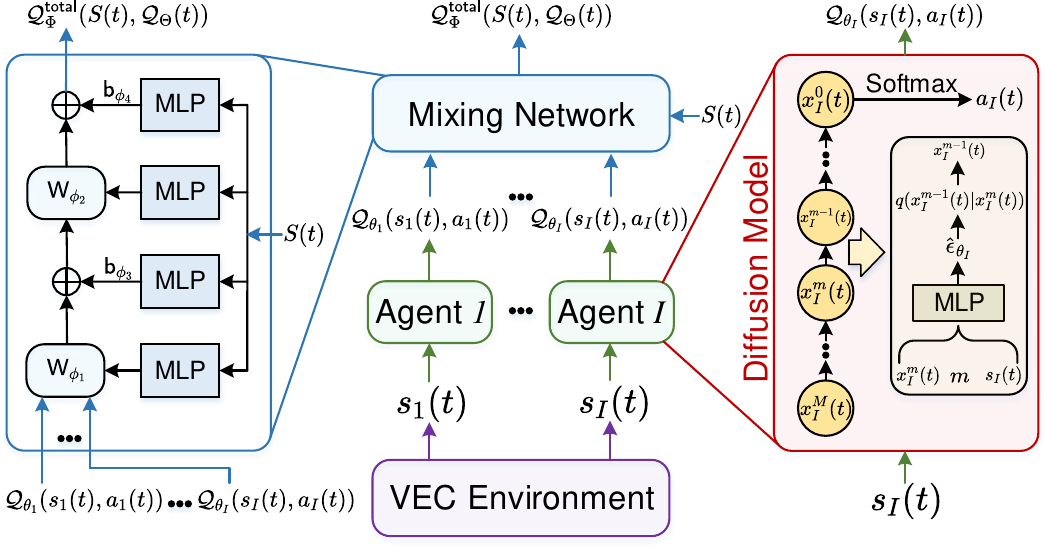}
\centering
\vspace{-1.5mm}
\caption{The overall architecture of the MAD2RL algorithm.}
\label{fig:D2RL}
\end{figure}

\vspace{-3mm}
\subsection{Algorithm Architecture}  \label{subsec:D2RL_architecture} 
The architecture of our proposed MAD2RL is depicted in Fig.~\ref{fig:D2RL}, utilizing the QMIX framework for network training. Specifically, each CV is regarded an agent that makes local action decisions based on its local observation. As a result, MAD2RL comprises $I$ online agent networks based on diffusion models, each responsible for evaluating the local $q$-values for each agent, and an online centralized mixing network that assesses the quality of the integrated decisions. Both online agent networks and the mixing network are paired with respective target networks, designed to address the issue of training instability. Additionally, a replay buffer is employed to diminish sample correlation through random sampling.
\vspace{-3mm}
\subsubsection{Diffusion Model-Based Agent Network} \label{subsubsec:actor_network} 
In MAD2RL, the core of the agent network $\mathcal{Q}_{\bm{\theta}_i}(\cdot)$ of agent $i$, parameterized by $\bm{\theta}_i$, is the diffusion model detailed in Sec.~\ref{sec:DDPM} (depicted in red in Fig.~\ref{fig:D2RL}). Additionally, a target agent network $\hat{\mathcal{Q}}_{\hat{\bm{\theta}_i}}(\cdot)$, parameterized by $\bm{\hat{\theta}}_i$, is leveraged to stabilize the learning process and improve learning efficiency, which has the same network structure as $\mathcal{Q}_{\bm{\theta}_i}(\cdot)$.
\vspace{-2mm}
\subsubsection{Mixing Network} The feed-forward mixing network $\mathcal{Q}^{\mathsf{total}}_{\bm{\Phi}}(\bm{S}(t), \bm{\mathcal{Q}}_{\bm{\Theta}}(t))$ takes the joint state $\bm{S}(t)$ and the joint action-value $\bm{\mathcal{Q}}_{\bm{\Theta}}(t)=\{\mathcal{Q}_{\bm{\theta}_1}(\bm{s}_1(t),a_1(t)),...,\mathcal{Q}_{\bm{\theta}_I}(\bm{s}_I(t),a_I(t)\}$ as inputs to evaluate the quality of the integrated decisions:
\vspace{-1.5mm}
\begin{align}\label{eq:QMIX}
   &\mathcal{Q}_{\bm{\Phi}}^{\mathsf{total}}(\bm{S}(t), \bm{\mathcal{Q}}_{\bm{\Theta}}(t))\nonumber\\
   &= \bm{\mathsf{W}}_{\bm{\phi}_2}(\bm{S}(t))\big[\bm{\mathsf{W}}_{\bm{\phi}_1}(\bm{S}(t))\bm{\mathcal{Q}}_{\bm{\Theta}}(t)+\bm{\mathsf{b}}_{\bm{\phi}_3}(\bm{S}(t))\big]\nonumber\\
   &+\bm{\mathsf{b}}_{\bm{\phi}_4}(\bm{S}(t)), \bm{\Phi}=\{\bm{\phi}_1,\bm{\phi}_2,\bm{\phi}_3,\bm{\phi}_4\}, \bm{\Theta} = \{\bm{\theta}_i\}_{i=1}^I,
\end{align}
where $\bm{\mathsf{W}}_{\bm{\phi}_1}(\bm{S}(t))$, $\bm{\mathsf{W}}_{\bm{\phi}_2}(\bm{S}(t))$, $\bm{\mathsf{b}}_{\bm{\phi}_3}(\bm{S}(t))$, and $\bm{\mathsf{b}}_{\bm{\phi}_4}(\bm{S}(t))$ represent the weights and biases provided by hypernetworks, parameterized by $\bm{\phi}_1$, $\bm{\phi}_2$, $\bm{\phi}_3$, and $\bm{\phi}_4$, respectively, with the joint state $\bm{S}(t)$ as input. Similarly, a target mixing network $\hat{\mathcal{Q}}_{\bm{\hat{\Phi}}}^{\mathsf{total}}(\cdot)$, parameterized by $\bm{\hat{\Phi}}$, is proposed with the same network structure as $\mathcal{Q}_{\bm{\Phi}}^{\mathsf{total}}(\cdot)$.
\vspace{-2mm}
\subsubsection{Replay Buffer} During the training process, a replay buffer $\mathcal{E}$ is employed to diminish sample correlation through random sampling. To simplify the presentation, we employ $\bm{S}$, $\bm{A}$, and $r$ to denote the joint state, joint action, and common reward in the current time slot, respectively, while $\bm{S}^{\prime}$, $\bm{A}^{\prime}$, and $r^{\prime}$ correspond to the joint state, joint action, and common reward in the subsequent time slot. Then, at each time slot, MAD2RL stores the transition tuple $<\bm{S},\bm{A},r,\bm{S}^{\prime}>$ into $\mathcal{E}$, where it awaits sampling for the development of network policies.
\vspace{-2mm}
\subsubsection{Policy Improvement} After the initial phase of exploration, we sample a random mini-batch of $E$ samples $\{(\bm{S}_e,\bm{A}_e,r_e,\bm{S}^{\prime}_e)\}_{e=1}^{E}$ from the replay buffer $\mathcal{E}$ to update agent networks and the mixing network simultaneously. Note that QMIX employs a centralized training approach, where it has access to all agents' observations and actions during the training phase. However, during execution, each agent operates independently based on its own local observation, making the approach scalable and practical for real-world applications where individual agents may not have access to complete global information. Specifically, we minimize the average Temporal Difference (TD) error between the target Q-value $\hat{y}_e$ and the Q-value of the mixing network $\mathcal{Q}^{\mathsf{total}}_{\bm{\Phi}}(\cdot)$, which is given by
\vspace{-1.5mm}
\begin{align}\label{eq:critic_TD}
    &\quad\quad \min_{\{\bm{\Phi},\bm{\Theta}\}} \frac{1}{E}\sum_{e=1}^E\Big[ \frac{1}{2}(\hat{y}_e-\mathcal{Q}^{\mathsf{total}}_{\bm{\Phi}}(\bm{S}_e,\mathcal{Q}_{\bm{\Theta}}))^2  \Big], \\
    &\text{s.t.}\ \hat{y}_e=r_e+\omega \hat{\mathcal{Q}}^{\mathsf{total}}_{\bm{\hat{\Phi}}}(\bm{S}^{\prime}_e,\hat{\mathcal{Q}}_{\hat{\bm{\Theta}}}), \ \forall e\in\{1,2,...,E\}.
    \vspace{-.5mm}
\end{align}
Here, $\omega$ denotes the discount factor for future rewards, and the target Q-value $\hat{y}_e$ is calculated through the target mixing network $\hat{\mathcal{Q}}^{\mathsf{total}}_{\bm{\hat{\Phi}}}(\cdot)$, where $\hat{\mathcal{Q}}_{\hat{\bm{\Theta}}} = \{\max_{a}\hat{\mathcal{Q}}_{\hat{\bm{\theta}}_1}(\bm{s}^{\prime}_1,a),...,\max_{a}\hat{\mathcal{Q}}_{\hat{\bm{\theta}}_I}(\bm{s}^{\prime}_I,a) \}$. Then, the estimation accuracy of both $\mathcal{Q}_{\bm{\theta}_i}(\cdot)$ and $\mathcal{Q}_{\bm{\Phi}}^{\mathsf{total}}(\cdot)$ will be improved through iteratively minimizing the loss in~\eqref{eq:critic_TD} using standard gradient descent-based optimizers, such as Adam~\cite{du2023diffusionbased}.

During the training phase of each time step in MAD2RL, the parameters of the target networks are only partially updated, ensuring that the changes in the value functions are smooth over time. Following this phase, the parameters of the target networks are softly updated towards those of the corresponding online networks as follows:
\vspace{-1.5mm}
\begin{align}
    &\hat{\bm{\theta}}_i\leftarrow \varepsilon\bm{\theta}_i+(1-\varepsilon)\hat{\bm{\theta}}_i, \ \forall i \in \mathcal{I},\label{eq:soft_updateactor}\\
    &\hat{\bm{\Phi}} \leftarrow \varepsilon\bm{\Phi}+(1-\varepsilon)\hat{\bm{\Phi}}, \label{eq:soft_updatecritic}
    \vspace{-.5mm}
\end{align}
where $\varepsilon \in (0,1]$ determines the target network update rate. A smaller value of $\varepsilon$ results in slower updates, which can stabilize the learning; however, it comes with the cost of a longer training time. By adjusting $\varepsilon$, the stability of the target networks and the learning speed are tuned.

\begin{algorithm} [!t]
  \SetAlgoLined
  \SetKwData{Left}{left}\SetKwData{This}{this}\SetKwData{Up}{up}
  \SetKwFunction{Union}{Union}\SetKwFunction{FindCompress}{FindCompress}
  \SetKwInOut{Input}{input}\SetKwInOut{Output}{output}

  Initialize agent networks $\mathcal{Q}_{\bm{\theta}_i}(\cdot)$ and the mixing network $\mathcal{Q}^{\mathsf{total}}_{\bm{\Phi}}(\cdot)$, the discount factor $\omega$, the maximum learning episode $H$, the maximum training steps $T$ per episode.
  
  Initialize the target agent networks $\{\bm{\hat{\theta}}_i\}_{i=1}^I\leftarrow \{\bm{\theta}_i\}_{i=1}^I$ and the mixing network $\bm{\hat{\Phi}}\leftarrow \bm{\Phi}$, the replay buffer $\mathcal{E}$. 
  \BlankLine
  \For{$episode=1$ \KwTo $H$}{
    Receive initial local state $\{\bm{s}_i(1)\}_{i=1}^I$ for each agent.
    
    \For{$t=1$ \KwTo $T$}{
        \emph{// The following loop is run on each agent:}
        
        \For{$i=1$ \KwTo $I$}{
            Initialize a distribution $\bm{x}^M_i(t)\sim\mathcal{N}(0,\mathbf{I})$.
      
            \For{$m=M$ \KwTo $1$} {
            Use a deep neural network to infer the noise $\hat{\bm{\epsilon}}_{\bm{\theta}_i}(\bm{x}^m_i(t),m,\bm{s}_i(t))$.
        
            Calculate the mean $\bm{\mu}^m_{\bm{\theta}_i}(\bm{x}^{m}_i(t),m,\bm{s}_i(t))$ and the distribution $q(\bm{x}^{m-1}_i(t)|\bm{x}^{m}_i(t))$ by~\eqref{eq:reverse_mean} and~\eqref{eq:reverse_distribution}, respectively. 

            Calculate the distribution $\bm{x}^{m-1}_i(t)$ using the reparameterization technique by~\eqref{eq:reverse_update}.
		  }
        Calculate the probability distribution $\mathcal{Q}_{\bm{\theta}_i}(\bm{s}_i(t))$ based on~\eqref{eq:softmax_distribution} and determine the local action $a_i(t)$ based on~\eqref{eq:MDP_action}.
      
        Determine the DNN partitioning decision $\varphi_{i}(t) = \lfloor \frac{a_i(t)}{J} \rfloor$ and task offloading decision $\xi_{i}(t) =  a_i(t) \mod J$, then obtain $\{f^{\mathsf{rsu}}_k\}_{k=1}^{K}$ by~\eqref{eq:nu} and~\eqref{eq:optnu}.
        
        Execute $\{\varphi_{i}(t), \xi_{i}(t), f^{\mathsf{rsu}}_k\}$ in the environment, receive the common reward $r(t)$, and transfer to next state $\bm{s}_i(t+1)$.}
      
      Store the transition $(\bm{S}(t),\bm{A}(t),r(t),\bm{S}(t+1))$ in the replay buffer $\mathcal{E}$.

      Randomly sample a batch of $E$ transitions $\{(\bm{S}_e,\bm{A}_e,r_e,\bm{S}^{\prime}_e)\}_{e=1}^E$ from replay buffer $\mathcal{E}$.
      
      Simultaneously update agent networks' parameters $\bm{\Theta} = \{\bm{\theta}_i\}_{i=1}^I$ and mixing network's parameters $\bm{\Phi}=\{\bm{\phi}_1,\bm{\phi}_2,\bm{\phi}_3,\bm{\phi}_4\}$ by minimizing~\eqref{eq:critic_TD} using the sampled batch of data.
      
      Update the target networks' parameters $\{\bm{\hat{\theta}}_i\}_{i=1}^I$ and $\bm{\hat{\Phi}}$ by~\eqref{eq:soft_updateactor} and~\eqref{eq:soft_updatecritic}, respectively.
    }   
  }
  \caption{MAD2RL Algorithm}\label{algo：D2RL}
\end{algorithm}\DecMargin{1em}

\vspace{-3mm}
\subsection{Algorithm Complexity}  \label{subsec:complexity}  
Algorithm~\ref{algo：D2RL} outlines the pseudocode of our proposed MAD2RL. According to the analysis in~\cite{9573404}, since the training process can be implemented on a server with sufficient computational resources, we mainly focus on the computational complexity of the execution process on the CVs. Additionally, various CVs deployed by different DRL agents run in parallel across the VEC network, so the overall complexity of the multi-agent system can be determined by the complexity of a single agent on an CV. Specifically, at time slot $t$, for CV/agent $i \in \mathcal{I}_k$, we consider the MLP in the diffusion model contains $F$ fully connected layers, then the computational complexity of online execution (i.e., the reverse process of the diffusion model) can be calculated as $\mathcal{O}(M(|\bm{s}_i(t)|f_0+\sum_{h=1}^{F}f_{h-1}f_h+f_F(L_k+1)J))$, where $f_h$ represents the number of neurons in hidden layer $h$. 

\vspace{-3mm}
\section{Performance Evaluation}  \label{sec:simulation}  
In this section, we first outline the parameter settings for simulations and subsequently evaluate the performance of our proposed MAD2RL by comparing it against three benchmarks.
\vspace{-3mm}
\subsection{Simulation Parameters}

\begin{figure}[htb]
    \centering

    \begin{subfigure}[b]{0.24\textwidth}  
        \includegraphics[width=\textwidth]{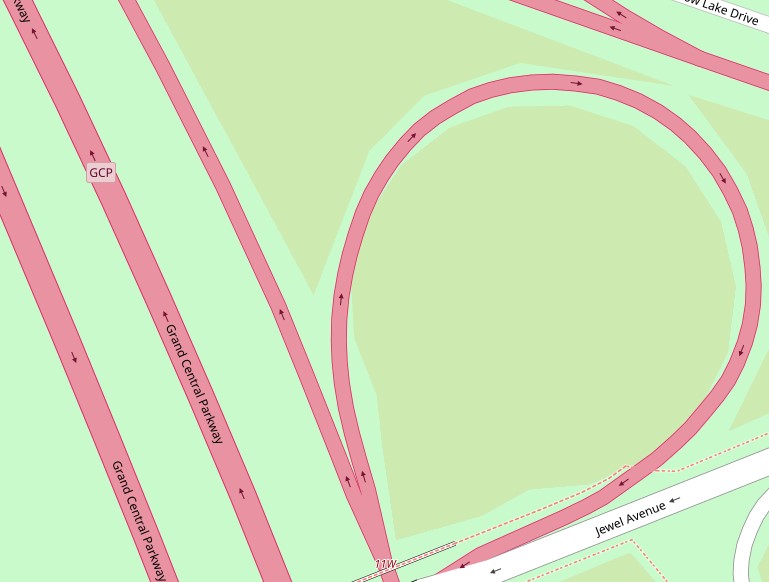}  
        \caption{Real-world traffic region.}
        \label{fig:real-world traffic}
    \end{subfigure}
    \hfill  
    \begin{subfigure}[b]{0.24\textwidth}
        \includegraphics[width=\textwidth]{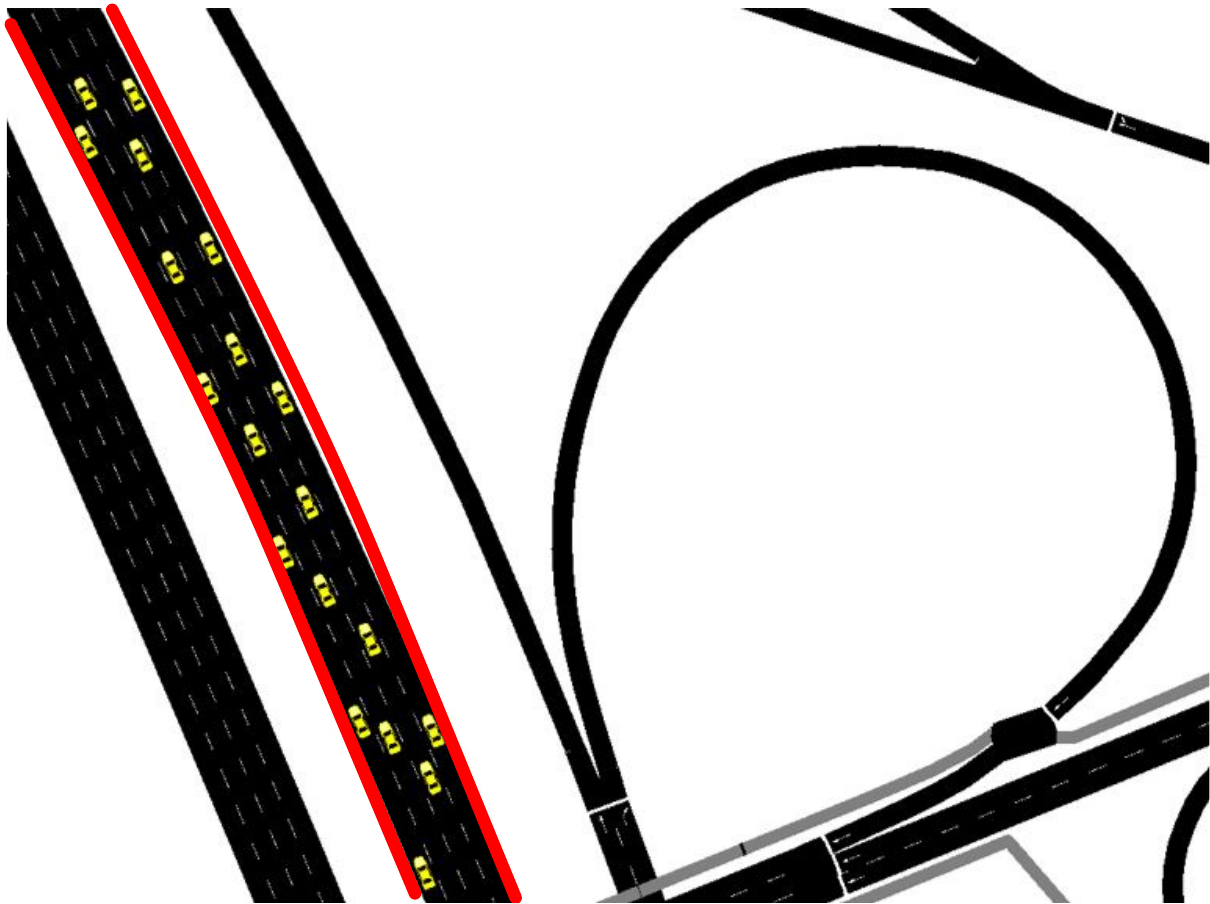}  
        \caption{Import moving vehicles.}
        \label{fig:moving_vehicles}
    \end{subfigure}

    \caption{Vehicular network visualization.}
    \label{fig:vehicular_networks}
    \vspace{-2mm}
\end{figure}

\emph{1) Vehicular Networks:} We consider a real-world traffic region in New York, USA as shown in Fig.~\ref{fig:real-world traffic} obtained from OpenStreetMap~\cite{4653466}. Subsequently, SUMO~\cite{8569938} is leveraged to import moving vehicles in an unidirectional highway (highlighted in red) and form a realistic vehicular network as shown in Fig.~\ref{fig:moving_vehicles}. We consider the bandwidth $\mathcal{B}=10$ (in MHz), and the noise power $\sigma^2 = -114$ (in dBm).

\emph{2) Parameters of CVs and SVs:} We assume that the computation capability of CVs and SVs is uniformly distributed in the range of [4, 6] (in GHz) and [6, 8] (in GHz)~\cite{10141674}, respectively. The computation capability of the RSU is 30 (in GHz). The transmit power of CV is chosen to be 23 (in dBm). Also, the distance between different CVs and SVs at different time slots are captured by SUMO~\cite{8569938}.  

\emph{3) DNN Model Types:} We adopt three classical DNN models, including AlexNet~\cite{10.1145/3065386}, ResNet18~\cite{7780459}, and VGG16~\cite{simonyan2015deep}, where the parameters of the convolution, pooling and fully-connected layers are extracted accordingly. The memory footprint $\varrho$ for a unit data is set to 4 (in Byte)~\cite{10141674}. The total number of time slots is set as $T=30$ (in second), and the duration of each time slot $\tau$ is 1 (in second).

\emph{4) Parameters of Neural Networks:} We implement MAD2RL using PyTorch 2.0 and Python 3.8.1 platforms. For the diffusion model of each agent, we configure the deep neural network to learn noise with 3 Fully Connected (FC) hidden layers, each consisting of 256 neurons. Additionally, we employ 2 FC hidden layers for each hypernetwork, each containing 64 neurons. The learning rate is set to $5 \times 10^{-4}$ for the Adam optimizer~\cite{du2023diffusionbased}. We set the maximum number of episodes $H$ to 1000, and the maximum steps (maximum time slots) per episode $T$ to 30. The reward discount factor and the target network update rate are 0.99 and 0.001, respectively.

\begin{figure}[t!]
    \centering
    \begin{subfigure}[b]{.38\textwidth}
        \includegraphics[width=\textwidth]{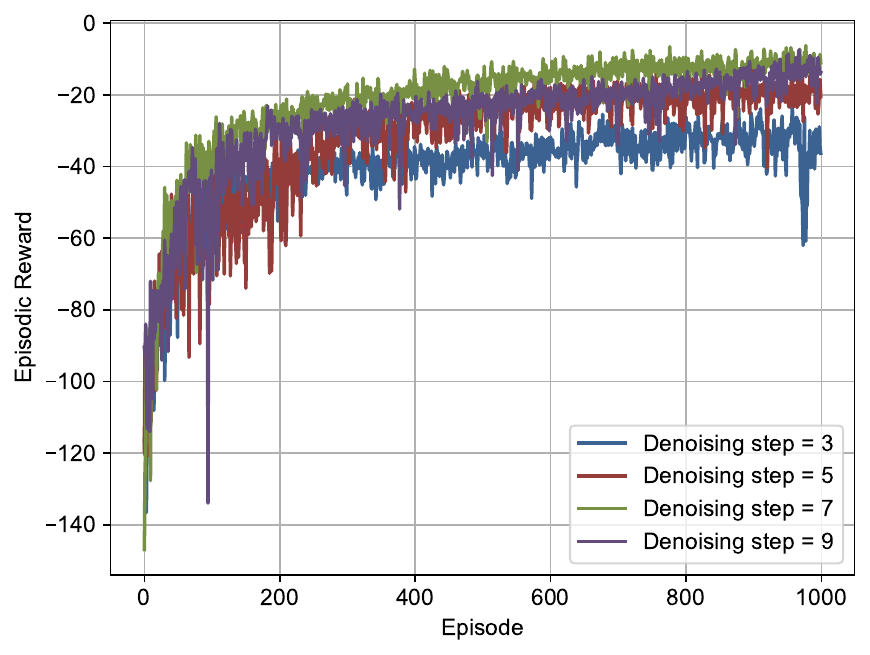}
        \vspace{-6.5mm}
        \caption{Denoising step impact on the reward.}
        \label{fig:D2RL_convergency_different_steps}
    \end{subfigure}
    \vspace{-3mm} 
    \begin{subfigure}[b]{.38\textwidth}
        \includegraphics[width=\textwidth]{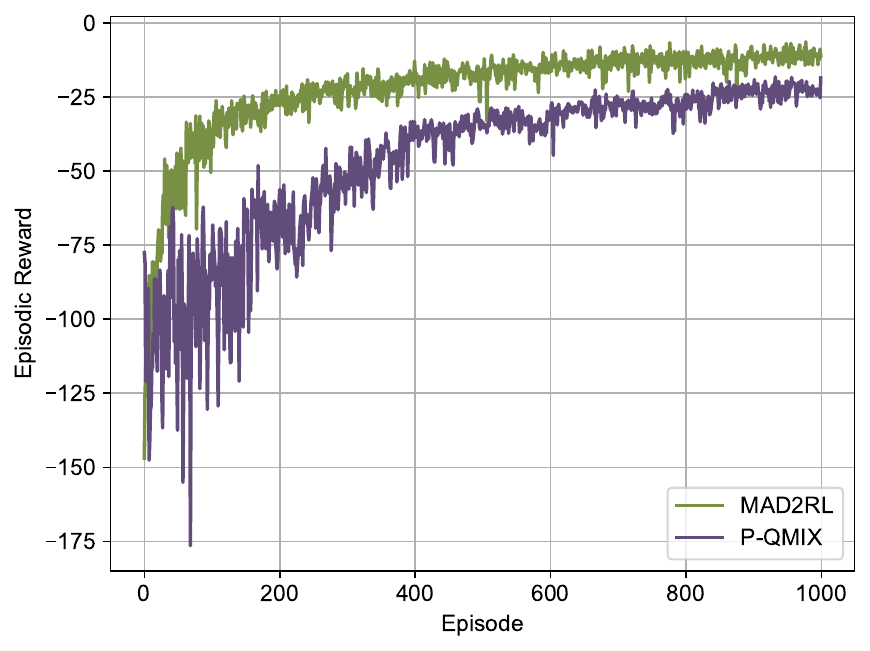}
        \vspace{-6.5mm}
        \caption{Comparison of reward curves of MAD2RL and P-QMIX.}
        \label{fig:D2RL_QMIX_convergency}
    \end{subfigure}
    \vspace{2mm}
    \caption{Convergence performance analysis.}
    \label{fig:test}
\end{figure}

\begin{figure*}[htbp]
    \centering
    \begin{subfigure}[b]{0.32\textwidth}  
        \includegraphics[width=\linewidth]{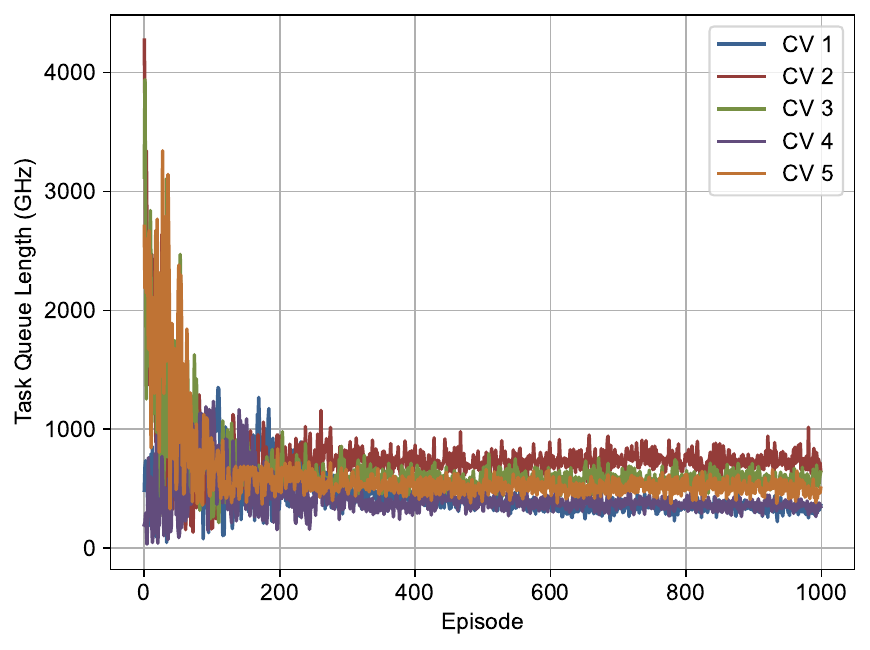}
        \vspace{-3.5mm}
        \caption{Task queue length of different CVs.}
        \label{fig:cv_queue_stability}
    \end{subfigure}
    \hfill
    \begin{subfigure}[b]{0.32\textwidth}  
        \includegraphics[width=\linewidth]{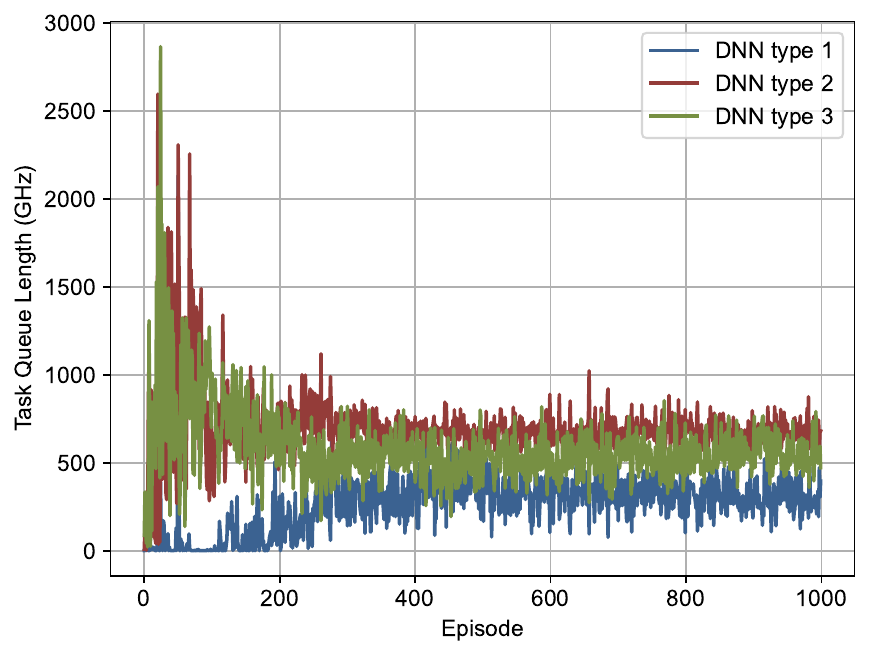}
         \vspace{-3.5mm}
        \caption{Task queue length of different DNNs.}
        \label{fig:dnn_queue_stability}
    \end{subfigure}
    \hfill
    \begin{subfigure}[b]{0.32\textwidth}  
        \includegraphics[width=\linewidth]{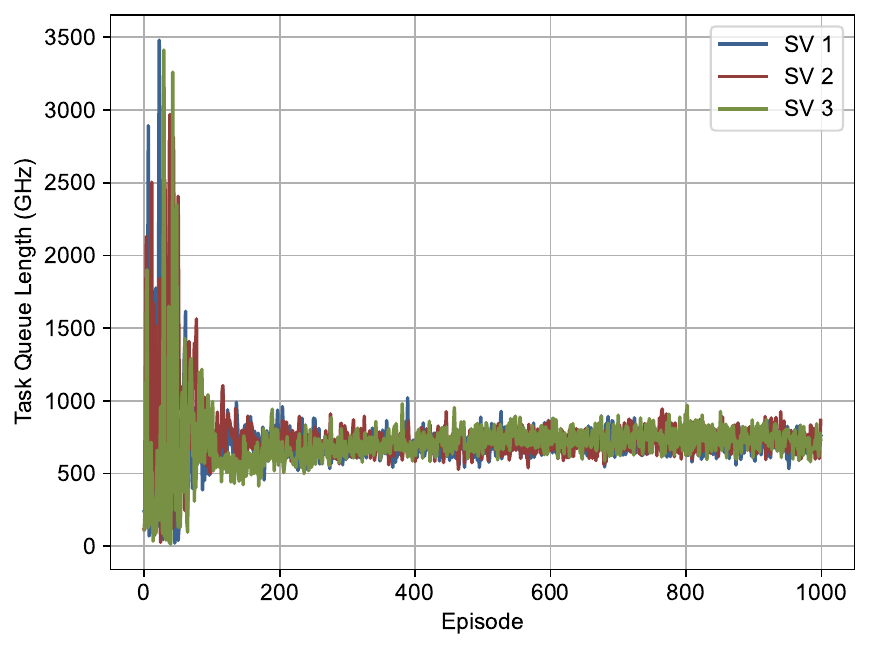}
         \vspace{-3.5mm}
        \caption{Task queue length of different SVs.}
        \label{fig:sv_queue_stability}
    \end{subfigure}
    \caption{Stability of the involved queues.}
    \label{fig:queue_stability}
    \vspace{-6.5mm}
\end{figure*}

\vspace{-3mm}
\subsection{Benchmark Solutions}

To study the performance of MAD2RL, we have proposed three benchmark solutions, including both DRL-based and heuristic-based methods as detailed below. Note that the computation resource allocation for these benchmark solutions is also determined by the convex techniques proposed in this work, following the acquisition of the corresponding DNN partitioning and task offloading decisions.

\begin{itemize}
	\item \emph{Pure QMIX (P-QMIX)~\cite{rashid2020monotonic}:} Contrasting with our proposed MAD2RL, P-QMIX utilizes an MLP-based agent network for each agent to make the optimal decisions on DNN partitioning and task offloading. This method is employed to highlight the substantial benefits of the diffusion model leveraged in this work.
    \item \emph{Genetic algorithm:} Drawing inspiration from~\cite{8451923}, at the beginning of each time slot, genetic algorithm generates several chromosomes. Each chromosome encompasses the DNN partitioning and task offloading decisions for all CVs, constituting the initial population. These chromosomes are then refined through a series of hybridizations and mutations until the maximum number of iterations is reached. Finally, the chromosome that yields the lowest value of~\eqref{eq:problem_2} is selected. 
	\item \emph{Greedy algorithm}: This is a rule-based algorithm, where at the start of each time slot, each CV chooses the DNN layer with the smallest transmission data size as its DNN partitioning decision. Subsequently, the CVs select the edge node with the shortest task queue as their task offloading decision. This approach establishes a lower-bound baseline for evaluating scheduling performance.
\end{itemize}

\vspace{-3mm}
\subsection{Simulation Results}

\subsubsection{Convergence Performance}
Fig.~\ref{fig:D2RL_convergency_different_steps} illustrates the episodic reward curves as training episodes proceed for our proposed MAD2RL under various denoising steps. This experiment was conducted with 7 CVs, 5 SVs, and $V=10$. We observe that the reward initially increases but then decreases with an increasing number of denoising steps. This occurs because as the number of denoising steps increases, the training process generally becomes more stable, reducing high-frequency oscillations and helping the diffusion model learn more general features. However, more denoising steps can lead to poorer performance. The reason is that too many denoising steps can cause the diffusion model to remove excessive noise, eliminating valuable details from the data. Additionally, the errors introduced in each denoising step can continuously accumulate, resulting in an increasing deviation between the final result and the real distribution, thereby compromising the output quality of the model. Consequently, we fix the denoising steps of MAD2RL to 7 when comparing it with the benchmark in subsequent analyses.

As shown in Fig.~\ref{fig:D2RL_QMIX_convergency}, we illustrate the convergence behavior of both our proposed MAD2RL and P-QMIX as the number of training episodes increases. This experiment was conducted with 7 CVs, 5 SVs, and $V=10$. Overall, our proposed MAD2RL is more stable and achieves a higher episodic reward during the training phase, demonstrating substantial benefits from the diffusion mode. This improvement is due to the generative capabilities of diffusion models, which significantly enhance action sample efficiency by progressively reducing noise through various denoising steps. Specifically, MAD2RL has achieved about 52\% improvement in the reward compared with P-QMIX.
\vspace{-2mm}
\subsubsection{Stability of the Involved Queues}
We evaluate the system stability achieved via the integration of Lyapunov technique in our methodology in Fig.~\ref{fig:queue_stability} by depicting the task queue length as the number of training episodes increases. This experiment was conducted with 5 CVs, 3 SVs, and $V=10$. Specifically, Fig.~\ref{fig:cv_queue_stability} illustrates the task queue stability of different CVs, demonstrated by the rapid convergence of the task queue length. Similarly, Fig.~\ref{fig:dnn_queue_stability} and Fig.~\ref{fig:sv_queue_stability} show the task queue stability for different DNN model types and SVs, respectively. Consequently, the convergence behavior of these time-evolving task queues indicates a stable task assignment, thereby guaranteeing the stable system operation considered in this work.

\subsubsection{Effect of the Number of Client Vehicles}
\begin{figure}[htb]
    \centering

    \begin{subfigure}[b]{0.24\textwidth}  
        \includegraphics[width=\textwidth]{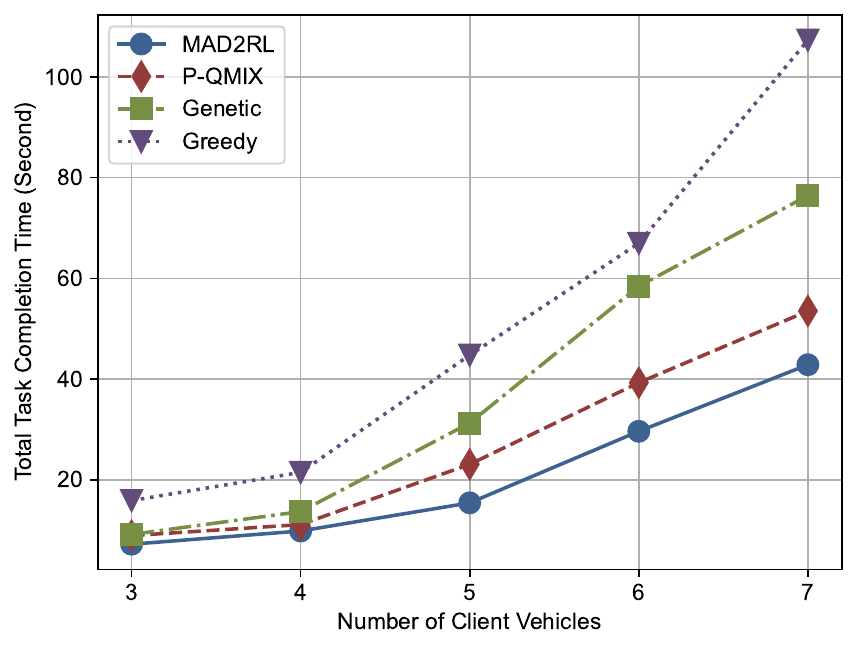}  
        \caption{Effect of the number of CVs.}
        \label{fig:CV_number}
    \end{subfigure}
    \hfill  
    \begin{subfigure}[b]{0.24\textwidth}
        \includegraphics[width=\textwidth]{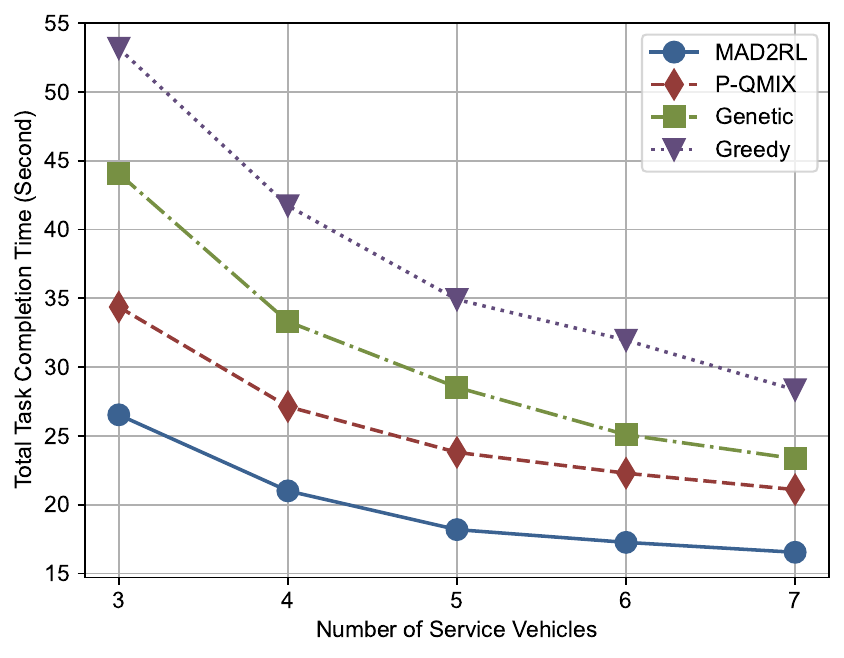}  
        \caption{Effect of the number of SVs.}
        \label{fig:SV_number}
    \end{subfigure}

    \caption{Performance evaluations upon considering different numbers of vehicles.}
    \label{fig:vehicular_networks}
    \vspace{-3.5mm}
\end{figure}

The results shown in Fig.~\ref{fig:CV_number} illustrate the effect of incrementally increasing the number of CVs from 3 to 7 on the overall task completion time. This experiment was conducted with 5 SVs, and $V=10$. We observe that as the number of CVs increases, the overall task completion time grows. The reason is that while the available computation resources of edge nodes remain unchanged, the number of DNN tasks requiring inference increases with the increase in CVs, leading to a longer task queue and thus an increase in task completion time. Overall, our proposed MAD2RL method outperforms other algorithms. Its performance is 54.59\% better than the greedy algorithm, 21.23\% better than the genetic algorithm, and 19.14\% better than P-QMIX at 3 CVs; it is 60.06\% better than the greedy algorithm, 43.94\% better than the genetic algorithm, and 19.93\% better than P-QMIX at 7 CVs.

\begin{figure}[htb]
    \centering

    \begin{subfigure}[b]{0.24\textwidth}  
        \includegraphics[width=\textwidth]{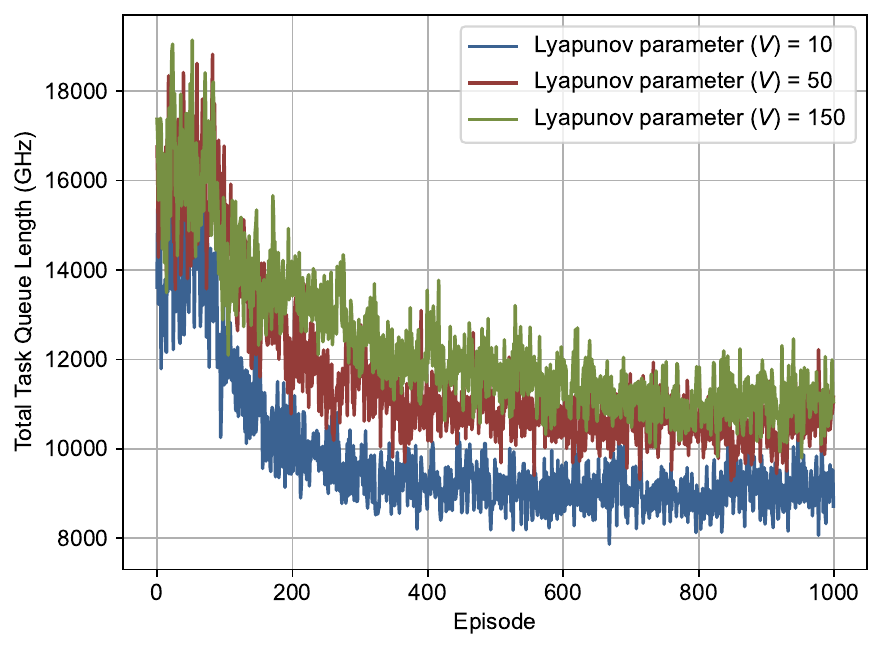}  
        \caption{Total task queue length.}
        \label{fig:total_task_queue}
    \end{subfigure}
    \hfill  
    \begin{subfigure}[b]{0.24\textwidth}
        \includegraphics[width=\textwidth]{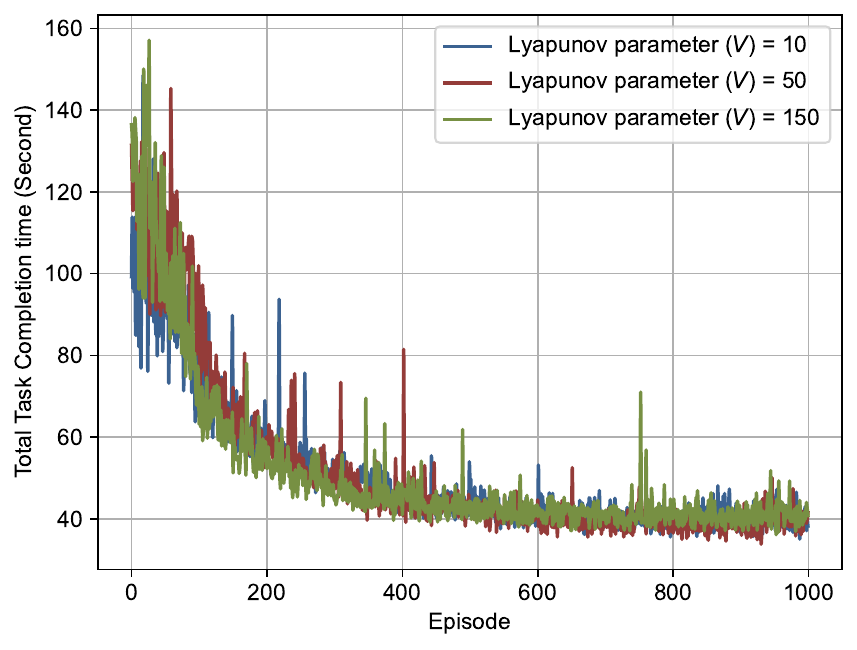}  
        \caption{Total task completion time.}
        \label{fig:total_task_completion_time}
    \end{subfigure}

    \caption{Effect of the Lyapunov control parameter.}
    \label{fig:Lyapunov_parameter}
     \vspace{-2.5mm}
\end{figure}

\vspace{-2mm}
\subsubsection{Effect of the Number of Service Vehicles}
Fig.~\ref{fig:SV_number} shows the effect of the number of SVs on the overall task completion time. This experiment was conducted with 5 CVs, and $V=10$. We observe that as the number of SVs increases, the task completion time decreases significantly at first, but then flattens out when the number of SVs exceeds that of CVs. The reason is that, initially, as the number of SVs increases, CVs tend to offload tasks to idle SVs to avoid backlogs in the task queues. However, when the number of SVs surpasses that of CVs, increasing the number of SVs leads to redundancy in computing resources, and the task completion time does not improve significantly thereafter. Overall, MAD2RL outperforms other algorithms. Its performance is 50.1\% better than the greedy algorithm, 39.79\% better than the genetic algorithm, and 22.8\% better than P-QMIX at 3 SVs; it is 41.65\% better than the greedy algorithm, 29.15\% better than the genetic algorithm, and 21.58\% better than P-QMIX at 7 SVs. 
\vspace{-2mm}
\subsubsection{Effect of the Lyapunov Control Parameter $V$}
In Fig.~\ref{fig:total_task_queue} and Fig.~\ref{fig:total_task_completion_time}, we further illustrate the impact of the Lyapunov control parameter $V$ in~\eqref{eq:Lyapunovdriftpluspenalty} on the performance of task queue stability and overall task completion time. This experiment was conducted with 7 CVs, and 5 SVs. Generally, as the weight of the Lyapunov control parameter $V$ increases (i.e., more emphasis is placed on task completion time), the average backlog in different task queues exhibits an increase with the control parameter, while the overall task completion time remains unchanged. Although the delay remains unchanged when a larger control parameter is employed, it becomes comparatively more prominent as the overhead for guaranteeing stability increases.

\vspace{-3mm}
\section{Conclusion and Future Works} \label{sec:conclusion}
In this paper, we addressed the problem of joint DNN partitioning, task offloading, and resource allocation in VEC as a dynamic long-term optimization. Our objective was to minimize the DNN-based task completion time while guaranteeing the system stability over time. To achieve this, we first employed a Lyapunov optimization technique to decouple the
original long-term optimization problem with stability constraints
into a per-slot deterministic problem. Afterwards, we proposed a
MAD2RL algorithm, incorporating the innovative use of a diffusion model, to determine the optimal DNN partitioning, and task offloading decisions. Furthermore, we integrated convex optimization techniques into MAD2RL as a
subroutine for allocating computation resources. Through numerical simulations, we demonstrated the superior performance of our proposed MAD2RL
algorithm compared to existing benchmark solutions. 

Future research could further explore the potential of utilizing an expert dataset which can be obtained offline through the brute-force search method, to conduct the forward process of diffusion models. This data set includes the optimal solutions for DNN partitioning and task offloading. Subsequently, a form of supervised learning could be applied to train the diffusion model to fit the action distribution generated by the reverse process and the expert data. Furthermore, investigations of competitions and cooperations among multiple RSUs for task acquisition is an enticing direction. 
\vspace{-3mm}
\bibliographystyle{IEEEtran}

\vspace{-14mm}
\clearpage
\begingroup

\appendices

\section{Proof of Lemma~\ref{lemma:upperbound}}\label{app:upperbound}

First, by squaring both sides of~\eqref{eq:local_queue},~\eqref{eq:rsu_queue} and~\eqref{eq:sv_queue}, we have
\vspace{-1mm}
\begin{align}\label{eq:localqueue_squa}
 Q_i^{\mathsf{loc}}(t+1)^2&= \Bigg[\max \Big\{Q_i^{\mathsf{loc}}(t)- f^{\mathsf{loc}}_i\tau\nonumber\\
 &+\mathbb{I}_{\{\varphi_{i}(t)\neq1\}}\Big(\sum_{l_{k}=1}^{\varphi_{i}(t)-1}B_{l_{k}}\Big),0 \Big\}\Bigg]^2, i \in \mathcal{I}_k.
\vspace{-.5mm}
\end{align}
\vspace{-3.5mm}
\begin{align}\label{eq:rsuqueue_squa}
    &Q^{\mathsf{rsu}}_k(t+1)^2=\Bigg[\max\Big\{Q_k^{\mathsf{rsu}}(t)-f^{\mathsf{rsu}}_k(t)\tau \nonumber\\
    &+\sum_{i \in \mathcal{I}_k}\xi_{i,1}(t)\mathbb{I}_{\{\varphi_{i}(t)\neq L_{k}+1\}}\Big(\sum_{l_{k}=\varphi_{i}(t)}^{L_{k}}B_{l_{k}}\Big),0 \Big\}\Bigg]^2,
    \vspace{-.5mm}
\end{align}
\vspace{-3.5mm}
\begin{align}\label{eq:svqueue_squa}
    &Q^{\mathsf{veh}}_j(t+1)^2=\Bigg[\max \Big\{Q^{\mathsf{veh}}_j(t)-f^{\mathsf{veh}}_j\tau \nonumber\\
    &+\sum_{k \in \mathcal{K}}\sum_{i \in \mathcal{I}_k}\xi_{i,j}(t)\mathbb{I}_{\{\varphi_{i}(t)\neq L_{k}+1\}}\Big(\sum_{l_{k}=\varphi_{i}(t)}^{L_{k}}B_{l_{k}}\Big),0 \Big\}\Bigg]^2.
    \vspace{-.5mm}
\end{align}

Then, by adopting inequality $\big[\max\{x,0\}\big]^2 \leq x^2$,~\eqref{eq:localqueue_squa},~\eqref{eq:rsuqueue_squa} and~\eqref{eq:svqueue_squa} can be transformed into the following equations, respectively,
\vspace{-1.5mm}
\begin{align}\label{eq:localqueue_ineq_1}
  &Q_i^{\mathsf{loc}}(t+1)^2\leq Q_i^{\mathsf{loc}}(t)^2\!+\!\! \Bigg[\mathbb{I}_{\{\varphi_{i}(t)\neq1\}}\Big(\sum_{l_{k}=1}^{\varphi_{i}(t)-1}B_{l_{k}}\Big)\!-\!f^{\mathsf{loc}}_i\tau\Bigg]^2\nonumber\\
  &+2Q_i^{\mathsf{loc}}(t)\Bigg[\mathbb{I}_{\{\varphi_{i}(t)\neq1\}}\Big(\sum_{l_{k}=1}^{\varphi_{i}(t)-1}B_{l_{k}}\Big)-f^{\mathsf{loc}}_i\tau\Bigg], i \in \mathcal{I}_k,
  \vspace{-.5mm}
\end{align}
\vspace{-4.5mm}
\begin{align}\label{eq:rsuqueue_ineq_1}
  &Q_k^{\mathsf{rsu}}(t+1)^2\leq Q_k^{\mathsf{rsu}}(t)^2\nonumber\\
  &+ \Bigg[\sum_{i \in \mathcal{I}_k}\xi_{i,1}(t)\mathbb{I}_{\{\varphi_{i}(t)\neq L_{k}+1\}}\Big(\sum_{l_{k}=\varphi_{i}(t)}^{L_{k}}B_{l_{k}}\Big)-f^{\mathsf{rsu}}_k(t)\tau\Bigg]^2 \nonumber\\
  &+\!2Q_k^{\mathsf{rsu}}(t)\!\Bigg[\!\sum_{i \in \mathcal{I}_k}\xi_{i,1}(t)\mathbb{I}_{\{\varphi_{i}(t)\neq L_{k}+1\}}\Big(\sum_{l_{k}=\varphi_{i}(t)}^{L_{k}}B_{l_{k}}\!\Big)\!-\!f^{\mathsf{rsu}}_k(t)\tau\!\Bigg],
  \vspace{-.5mm}
\end{align}
\vspace{-4.5mm}
\begin{align}\label{eq:svqueue_ineq_1}
&Q_j^{\mathsf{veh}}(t+1)^2\leq Q_j^{\mathsf{veh}}(t)^2\nonumber\\
  &+ \Bigg[\sum_{k \in \mathcal{K}}\sum_{i \in \mathcal{I}_k}\xi_{i,j}(t)\mathbb{I}_{\{\varphi_{i}(t)\neq L_{k}+1\}}\Big(\sum_{l_{k}=\varphi_{i}(t)}^{L_{k}}B_{l_{k}}\Big)-f^{\mathsf{veh}}_j\tau\Bigg]^2 \nonumber\\
  &+\!2Q_j^{\mathsf{veh}}(t)\!\Bigg[\!\sum_{k \in \mathcal{K}}\sum_{i \in \mathcal{I}_k}\!\xi_{i,j}(t)\mathbb{I}_{\{\varphi_{i}(t)\neq L_{k}+1\}}\!\Big(\!\sum_{l_{k}=\varphi_{i}(t)}^{L_{k}}B_{l_{k}}\!\Big)\!\!-\!f^{\mathsf{veh}}_j\tau\!\Bigg].
  \vspace{-.5mm}
\end{align}

Subsequently, based on inequality $(x-y)^2\leq x^2+y^2, \forall x,y \geq0$, according to~\eqref{eq:Lyapunovfunc} and~\eqref{eq:Lyapunovdrift}, we obtain
\vspace{-1.5mm}
\begin{align}\label{eq:Lyapunovdrift_ineq}
  &\Delta(\bm{Q}(t))= \mathbb{E}\Bigg[\frac{1}{2}\sum_{i \in\mathcal{I}}\Big[Q^{\mathsf{loc}}_i(t+1)^2 -Q^{\mathsf{loc}}_i(t)^2\Big]\nonumber\\
  &+ \frac{1}{2}\sum_{k\in \mathcal{K}}\Big[ Q^{\mathsf{rsu}}_k(t+1)^2-Q^{\mathsf{rsu}}_k(t)^2\Big]\nonumber\\
  &+ \frac{1}{2}\sum_{j\in \mathcal{J}\setminus\{1\}}\Big[Q^{\mathsf{veh}}_j(t+1)^2 -Q^{\mathsf{veh}}_j(t)^2\Big]\ \Big|\ \bm{Q}(t) \Bigg]\nonumber\\
  &\leq \mathbb{E}\Bigg[ \sum_{i\in \mathcal{I}}Q_i^{\mathsf{loc}}(t)\Bigg[\mathbb{I}_{\{\varphi_{i}(t)\neq1\}}\Big(\sum_{l_{k}=1}^{\varphi_{i}(t)-1}B_{l_{k}}\Big)-f^{\mathsf{loc}}_i\tau\Bigg]\nonumber\\
  &+\frac{1}{2}\sum_{i\in \mathcal{I}}\Big[\mathbb{I}_{\{\varphi_{i}(t)\neq1\}}\Big(\sum_{l_{k}=1}^{\varphi_{i}(t)-1}B_{l_{k}}\Big)\Big]^2+\frac{1}{2}\sum_{i\in \mathcal{I}}(f^{\mathsf{loc}}_i\tau)^2 \nonumber\\  
  &+\!\!\sum_{k \in\mathcal{K}}Q_k^{\mathsf{rsu}}(t)\!\Bigg[\!\sum_{i \in \mathcal{I}_k}\xi_{i,1}(t)\mathbb{I}_{\{\varphi_{i}(t)\neq L_{k}+1\}}\Big(\sum_{l_{k}=\varphi_{i}(t)}^{L_{k}}\!\!\!\!B_{l_{k}}\!\Big)\!-\!f^{\mathsf{rsu}}_k(t)\tau\!\Bigg]\nonumber\\
  &+\frac{1}{2}\sum_{k \in\mathcal{K}}\Big[ \sum_{i \in \mathcal{I}_k}\xi_{i,1}(t)\mathbb{I}_{\{\varphi_{i}(t)\neq L_{k}+1\}}\Big(\sum_{l_{k}=\varphi_{i}(t)}^{L_{k}}B_{l_{k}}\Big)\Big]^2 \nonumber\\
  &+ \frac{1}{2}\sum_{k \in\mathcal{K}}(f^{\mathsf{rsu}}_k(t)\tau)^2\nonumber\\
  &+\!\!\!\!\sum_{j\in \mathcal{J}\setminus\{1\}}\!\!\!\!\!\!Q_j^{\mathsf{veh}}(t)\!\Bigg[\!\sum_{k \in \mathcal{K}}\sum_{i \in \mathcal{I}_k}\!\xi_{i,j}(t)\mathbb{I}_{\{\varphi_{i}(t)\neq L_{k}+1\}}\!\Big(\!\sum_{l_{k}=\varphi_{i}(t)}^{L_{k}}\!\!\!\!\!B_{l_{k}}\!\Big)\!\!-\!\!f^{\mathsf{veh}}_j\tau\!\Bigg] \nonumber\\
  &+\frac{1}{2}\sum_{j\in \mathcal{J}\setminus\{1\}}\Big[\sum_{k \in \mathcal{K}}\sum_{i \in \mathcal{I}_k}\!\xi_{i,j}(t)\mathbb{I}_{\{\varphi_{i}(t)\neq L_{k}+1\}}\!\Big(\!\sum_{l_{k}=\varphi_{i}(t)}^{L_{k}}B_{l_{k}}\!\Big) \Big]^2 \nonumber\\
  &+\frac{1}{2}\sum_{j\in \mathcal{J}\setminus\{1\}}(f^{\mathsf{veh}}_j\tau)^2      \Big|\bm{Q}(t)\Bigg].
  \vspace{-.5mm}
\end{align}

Finally, after combining the elements independent of the task queues into a constant $\chi$ through giving the decisions on task offloading $\xi_{i,j}(t)$ and DNN partitioning $\varphi_i(t)$, the proof of the Lemma~\ref{lemma:upperbound} is obtained.

\end{document}